\documentclass[letterpaper, 10 pt, conference]{ieeeconf}  %

\IEEEoverridecommandlockouts                              %

\usepackage{url}

\usepackage{times}
\usepackage{epsfig}
\usepackage{graphicx}
\usepackage{amsmath}
\usepackage{amssymb}
\usepackage{multicol}
\usepackage{wrapfig}
\usepackage{lscape}
\usepackage{caption}
\usepackage{booktabs}
\usepackage{tabularx}
\usepackage{array}

\usepackage{algorithm2e}

\usepackage{bm}
\usepackage{bbm}

\usepackage{color}
\usepackage{xcolor}

\usepackage{soul}

\newcommand{\ourmethod}{Universal Visual Decomposer\xspace}
\newcommand{\acronym}{UVD\xspace}
\newcommand{\norm}[1]{\left\lVert#1\right\rVert}

\newcommand{\ie}{\textit{i.e.}\xspace}
\newcommand{\eg}{\textit{e.g.}\xspace}

\newcommand{\indom}{\textsc{InD}\xspace}
\newcommand{\outdom}{\textsc{OoD}\xspace}

\usepackage[noadjust]{cite}

\definecolor{snsgray}{RGB}{179, 179, 179}
\definecolor{snsorange}{RGB}{252, 141, 98}
\definecolor{snsblue}{RGB}{141, 160, 203}

\definecolor{coolgrey}{RGB}{157,157,157}
\definecolor{lightgrey}{RGB}{235,238,238}
\definecolor{lightteal}{RGB}{198,211,222}
\definecolor{cyan}{RGB}{136, 204, 238}
\definecolor{teal}{RGB}{68, 170, 153}
\definecolor{sand}{RGB}{221, 204, 119}
\definecolor{rose}{RGB}{204, 102, 119}
\definecolor{red}{RGB}{250, 94, 91}
\definecolor{orange}{RGB}{255, 200, 63}
\definecolor{yellow}{RGB}{254, 239, 109}

\definecolor{darkgreen}{rgb}{0.09, 0.45, 0.27}

\usepackage{xcolor}
\definecolor{codegreen}{rgb}{0,0.6,0}
\definecolor{codegray}{rgb}{0.5,0.5,0.5}
\definecolor{codepink}{RGB}{252, 142, 172}
\definecolor{codepurple}{rgb}{0.58,0,0.82}
\definecolor{backcolour}{RGB}{250,250,250}
\usepackage{listings}
\lstdefinestyle{mystyle}{
    backgroundcolor=\color{backcolour},   
    commentstyle=\color{codegreen},
    keywordstyle=\color{blue},
    numberstyle=\tiny\color{codegray},
    stringstyle=\color{codepurple},
    basicstyle=\fontfamily{\ttdefault}\footnotesize,
    breakatwhitespace=false,         
    breaklines=true,                 
    captionpos=b,                    
    keepspaces=true,    
    frame=single,
    numbersep=5pt,                  
    showspaces=false,                
    showstringspaces=false,
    showtabs=false,                  
    tabsize=2,
    classoffset=1, %
    otherkeywords={range},
    keywordstyle=\color{violet},
    classoffset=0,
}

\usepackage{multirow, makecell}
\usepackage{multicol}
\newcommand{\bestscore}[1]{\textcolor{darkgreen}{\textbf{#1}}}

\newcommand{\citep}[1]{\cite{#1}}
\newcommand{\citet}[1]{\cite{#1}}
\usepackage{afterpage}
\usepackage{dblfloatfix}

\usepackage[pagebackref=true,breaklinks=true,letterpaper=true,colorlinks,bookmarks=false]{hyperref}

\usepackage[colorinlistoftodos]{todonotes}

\title{\LARGE \bf
Universal Visual Decomposer: Long-Horizon Manipulation Made Easy
}

\author{Zichen Zhang$^{*1}$, Yunshuang Li$^{*2}$, Osbert Bastani$^{2}$, Abhishek Gupta$^{3}$,\\ 
Dinesh Jayaraman$^{2}$, Yecheng Jason Ma$^{\dag2}$, Luca Weihs$^{\dag1}$%
\\ 
\href{https://zcczhang.github.io/UVD/}{
{\texttt{zcczhang.github.io/UVD}}
}%
\thanks{$^*$Equal Contribution. $^\dag$Equal Advising}%
\thanks{$^1$Allen Institute for AI, $^2$University of Pennsylvania, $^3$University of Washington}%
}

\begin{document}

\maketitle
\thispagestyle{empty}
\pagestyle{empty}

\begin{abstract}
Real-world robotic tasks stretch over extended horizons and encompass multiple stages. Learning long-horizon manipulation tasks, however, is a long-standing challenge, and demands decomposing the overarching task into several manageable subtasks to facilitate policy learning and generalization to unseen tasks. Prior task decomposition methods require task-specific knowledge, are computationally intensive, and cannot readily be applied to new tasks. To address these shortcomings, we propose \ourmethod (\acronym), an off-the-shelf task decomposition method for visual long-horizon manipulation using pre-trained visual representations designed for robotic control. At a high level, \acronym discovers subgoals by detecting phase shifts in the embedding space of the pre-trained representation. Operating purely on visual demonstrations without auxiliary information, \acronym can effectively extract visual subgoals embedded in the videos, while incurring zero additional training cost on top of standard visuomotor policy training. Goal-conditioned policies learned with \acronym-discovered subgoals exhibit significantly improved compositional generalization at test time to unseen tasks. Furthermore, \acronym-discovered subgoals can be used to construct goal-based reward shaping that jump-starts temporally extended exploration for reinforcement learning. We extensively evaluate \acronym on both simulation and real-world tasks, and in all cases, \acronym substantially outperforms baselines across imitation and reinforcement learning settings on in-domain and out-of-domain task sequences alike, validating the clear advantage of automated visual task decomposition within the simple, compact \acronym framework.

\end{abstract}

\section{Introduction}\label{sec:intro}

Real-world household tasks, such as cooking and tidying, often stretch over extended horizons and encompass multiple stages. In order for robots to be deployed in realistic environments, they must possess the capability to learn and perform long-horizon manipulation tasks from visual observations. Learning vision-based complex skills over long timescales, however, is challenging due to the problem of compounding errors, the vastness of the action and observation spaces, and the difficulty in providing meaningful learning signals for each step of the task. 

Given these challenges, it is necessary to \textit{decompose} a long-horizon task into several smaller subtasks to make learning manageable. Beyond improving the efficiency of learning, task decomposition facilitates learning reusable skills, promotes data-sharing across different trajectories, and further enables compositional generalization to unseen sequences of the learned subtasks.
Despite its usefulness, task decomposition is difficult to perform in practice, and most existing approaches require strong assumptions about tasks, datasets, or robotic platforms~\citep{nair2019hierarchical, pertsch2020long, mandlekar2020learning, fang2022planning, fang2023generalization,huang2019neural, xu2018neural, shiarlis2018taco,borja2022affordance, james2022q, shridhar2023perceiver, shi2023waypoint}.
These methods cannot be used in common settings where the agent only has access to video demonstrations of desired behavior on their robotic hardware and little else, motivating the need for an off-the-shelf approach that can readily decompose \emph{any} visual demonstration out-of-the-box.

\begin{figure}[t]
    \centering
    \includegraphics[width=1\linewidth]{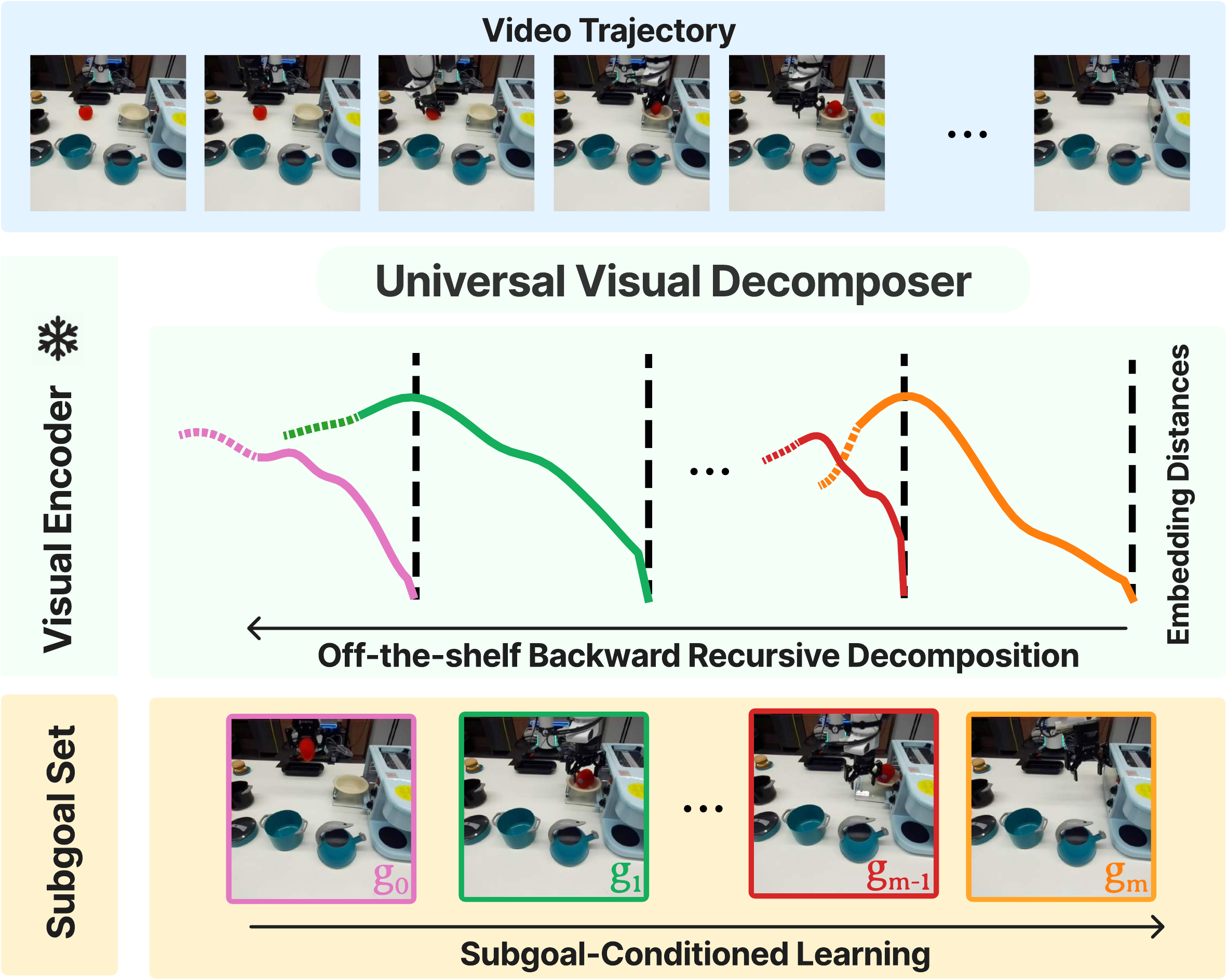}
    \vspace{-1em}
    \caption{{\textbf{\ourmethod} uses off-the-shelf pre-trained visual representations to find subgoals from video demonstrations by recursively computing embedding distances from the target goal and setting the first plateau as the new target goal.}}
    \label{fig:teaser}
    \vspace{-1em}
\end{figure}

In order to decompose any long-horizon task using vision, general knowledge about visual task progression that can discern embedded subtasks in long, unsegmented task videos must be acquired.  
 In this work, we propose \ourmethod (\acronym), an off-the-shelf unsupervised subgoal decomposition method that re-purposes state-of-the-art pre-trained visual representations~\citep{parisi2022unsurprising, nair2022r3m, khandelwal2022simple, xiao2022masked, ma2022vip, majumdar2023we, ma2023liv} for automated task segmentation. To motivate our approach, we observe that several pre-trained visual representations, such as VIP~\citep{ma2022vip} and R3M~\citep{nair2022r3m}, are trained to capture temporal task progress on diverse, short videos of humans accomplishing goal-directed behavior~\citep{damen2018scaling, grauman2022ego4d}. These representations have acquired well-behaved embedding distances that can progress near \textit{monotonically} along video frames that depict short-horizon, atomic skills.
 Our key insight is that when applied to long videos consisting of several subtasks, their training on short atomic tasks makes these representations no longer informative about subtask membership. That is, they are not trained to capture whether an earlier frame, which may very well belong to a different subtask, is making progress towards a subtask that appears later in the video, even if the subtasks are related to one another. As a consequence, when the robot task is extended, 
 the embedding distances will \textit{deviate} from monotonicity and exhibit plateaus around frames that correspond to phase shifts 
 in the overall task; this provides an unsupervised signal for detecting when subtasks have taken place in the original long, unsegmented task video. \acronym instantiates this insight and proposes an out-of-the-box subgoal discovery procedure that can iteratively extract subgoals using the embedding distance information from the end to the beginning; notably, \acronym does not require any domain-specific knowledge or incur additional training cost on top of standard visuomotor policy training. Given its off-the-shelf nature, \acronym can be readily applied to a variety of unseen robot domains. See Fig.~\ref{fig:teaser} for a conceptual overview of our approach.

We apply \acronym to long-horizon, multi-stage visual manipulation tasks in both simulation and real-world environments. Across these tasks, \acronym consistently outputs semantically meaningful subgoals which are used for policy training and evaluation. We consider both in-domain (\indom) and out-of-domain (\outdom) task evaluations. In \indom evaluation, the agent is evaluated on long-horizon tasks for which it has been explicitly trained whereas in \outdom evaluation, the agent is evaluated to generalize to new tasks unseen during training.
Using \acronym-discovered subgoals, we demonstrate substantial policy improvements across these evaluation settings.
Firstly, when training agents with reinforcement learning (RL), we show that \acronym-subgoals can be used to perform \textit{reward shaping} for each of the intermediate subtasks. Using this approach, we demonstrate that the resulting rewards can successfully guide a vision-based reinforcement learning agent to learn long-horizon tasks in the FrankaKitchen~\citep{gupta2019relay} environment. Secondly, when training agents with imitation learning (IL), by virtue of discovering semantically meaning subgoals, our policies can \textit{compositionally generalize} to \outdom task sequences unseen during training; this capability greatly reduces the burden of manual data collection for every desired task. 
Finally, in \indom evaluation, we also demonstrate performance improvement on several real-world multi-stage tasks that stretch over several hundred timesteps and exhibit sequential dependency among the subtasks.

In summary, our contributions include:
\begin{enumerate}
    \item \ourmethod (\acronym), an off-the-shelf visual decomposition method for long-horizon manipulation using pre-trained visual representations. 
    \item A reward shaping method for long-horizon visual reinforcement learning using \acronym-discovered subgoals.
    \item Extensive experiments demonstrating \acronym's effectiveness in improving policy performance on \indom and \outdom evaluations across several simulation and real-robot tasks.
\end{enumerate}

\section{Related Work}

Learning long-horizon skills has been a long standing challenge in robotic manipulation~\citep{gupta2019relay, mandlekar2020learning, lee2021ikea, heo2023furniturebench}. Hierarchical reinforcement learning~\citep{sutton1999between,shiarlis2018taco,bacon2017option,nachum2018near,nachum2018data,co2018self,bagaria2019option,gupta2019relay,eysenbach2019search,nasiriany2019planning,chane2021goal,zhang2021world} enables temporally extended exploration by discovering subskills and planning over them. However, these algorithms learn subskills and overall policies from scratch, which is computationally expensive and less suitable for real-world robotics use cases. 

When provided with task demonstrations, there are many prior efforts on using subgoal decomposition as a means to break up the long task in order to provide intermediate learning signals and to mitigate compounding action errors. These prior decomposition strategies, however, require task-specific knowledge and cannot be easily applied to new tasks. For example, several approaches use the robot's proprioceptive data within the task demonstrations~\citep{borja2022affordance, james2022q, shridhar2023perceiver, shi2023waypoint} or explicit knowledge about subtask structure~\citep{huang2019neural, xu2018neural} to guide decomposition; this limits the types of tasks that can be solved and precludes learning from observed videos. Other works learn latent generative models over subgoals~\citep{jayaraman2019time, nair2019hierarchical, pertsch2020long, mandlekar2020learning, fang2022planning, fang2023generalization}, but demand compute-intensive training on large datasets that cover diverse behavior.

To the best of our knowledge, \ourmethod is the first ``off-the-shelf'' visual task decomposition method that does not require any task-specific knowledge or training. In addition, it demonstrates a novel use case of pre-trained visual representations. While some prior works have considered using pre-trained visual representations to generate rewards~\citep{sermanet2018time, ma2022vip}, we are the first to demonstrate that they can also be re-purposed to perform hierarchical decomposition; furthermore, this capability can be combined with the reward specification capability to solve long-horizon tasks using visual reinforcement learning.

\section{Problem Setting}\label{sec:problem_setting}

\noindent \textbf{Unsupervised Subgoal Discovery (USD).} Our goal is to derive a general-purpose subgoal decomposition method that can operate purely from visual inputs on a \textit{per-trajectory} basis. That is, given a full-task demonstration $\tau = (o_0,..., o_T)$,
\begin{equation}
    \label{eq:usd}
    \texttt{USD}((o_0,...,o_T)) \rightarrow \tau_{goal}:=(g_0,...,g_m),
\end{equation}
where $(g_0,...,g_m)$ are the subset of $\tau$ that are selected as subgoals; $m$ may vary across trajectories.

\noindent \textbf{Policy Learning.} We provide demonstrations $\mathcal{D} := \{\tau\}_{i=1}^n$ for the learning tasks; in the reinforcement learning setting, we assume that there is one task and have $n=1$ to specify the overall task to be achieved. The evaluation tasks can be both in-domain (\indom), the ground-truth sequences of tasks captured in $\mathcal{D}$, or out-of-domain (\outdom), consisting of unseen combinations of the subtasks in $\mathcal{D}$. 

We assume access to a pre-trained visual representation $\phi : \mathbb{R}^{H \times W \times 3} \rightarrow \mathbb{R}^K$ that maps RGB images to a $K$-dimensional embedding space. Given $\phi$ and $\mathcal{D}$, our goal is to learn a goal-conditioned policy $\pi: \mathbb{R}^K {\times} \mathbb{R}^K {\rightarrow} \Delta(A)$ that outputs an action based on the embedded observation and goal, $a \sim \pi(\phi(o), \phi(g))$. In the RL setting, the agent 
is not provided with reward information, so the agent must also construct rewards using $\phi$ and $\mathcal{D}$. 

\noindent \textbf{Policy Evaluation.} For \outdom eval{.}, we provide one demonstration $\tau$ specifying the subtask sequence to be performed. 

\section{Method}\label{sec:method}
We first present \ourmethod, the core algorithm that powers our off-the-shelf subgoal discovery approach. Then, we discuss various ways we perform policy training as well as goal selection during policy inference. 

\subsection{\ourmethod}\label{subsec:ourmethod}

Given an unlabeled video demonstration $\tau =(o_0,...,o_T)$, how might we discover useful subgoals? The key intuition of Universal Visual Decomposer is that, conditioned on a goal frame $o_t$, some $n$ frames $(o_{t-n},..., o_{t-1})$ preceding it must visually approach the goal frame; once we discover the first frame ($o_{t-n}$) in this goal-reaching sequence, the frame that precedes it ($o_{t-n-1}$) is then another subgoal. From $o_{t-n-1}$, the same procedure can be carried out \textit{recursively} until we reach $o_0$.  There are two central questions to address: (1) how to discover the first subgoal (last in terms of timestamp), and (2) how to determine the stopping point for the current subgoal and declare a new frame as the new subgoal. 

The first question is simple to resolve by observing that in a demonstration, the last frame $o_T$ is naturally a goal. Now, conditioned on a subgoal $o_t$, we attempt to extract the first frame $o_{t-n}$ in the sub-sequence of frames that depicts visual task progression to $o_t$. To discover this first frame, we exploit the fact that several state-of-the-art pre-trained visual representations for robot control~\citep{nair2022r3m, ma2022vip, ma2023liv} are trained to capture temporal progress within short videos depicting a single solved task; these representations can effectively produce embedding distances that exhibit \textit{monotone} trend over a short goal-reaching video sequence $\tau=(o_{t-n},...,o_t)$: \
\begin{equation}
    \label{eq:monotonicity}
    d_\phi(o_s;o_t) \geq  d_\phi(o_{s+1};o_t), \forall s\in\{t-n, \ldots, t-1\},
\end{equation}
where $d_\phi$ is a distance function in the $\phi$-representation space; in this work, we set $d_\phi(o;o') := \norm{\phi(o)-\phi(o')}_2$ because several state-of-the-art pre-trained representations use the $L_2$ distance as their embedding metric for learning.
Given this, we set the previous subgoal to be the temporally closest observation to $o_t$ for which this monotonicity condition fails:
\begin{equation}
\label{eq:subgoal-criterion}
    o_{t-n-1} :=\arg \max_{o_h} d_\phi(o_h;o_t) < d_\phi(o_{h+1};o_t), h < t\ .
\end{equation}
The intuition is that a preceding frame that belongs to the same subtask (i.e., visually apparent that it is progressing towards $o_t$) should have a higher embedding distance than the succeeding frame if the embedding distance indeed captures temporal progression. As a result, a deviation from the monotonicity indicates that the preceding frame may not exhibit a clear relation to the current subgoal, and instead be a subgoal itself. 
Now, $o_{t-n-1}$ becomes the new subgoal, and we apply~\eqref{eq:subgoal-criterion} recursively until the full sequence $\tau$ is exhausted. For instance, in Figure~\ref{fig:teaser}, conditioned on the last frame, $g_3$ is the first preceding frame that produces an inflection point in the embedding distances and hence selected as a subgoal; then, conditioned on $g_3$, $g_2$ is selected, and so on; see Alg.~\ref{algo:decomp} {for high-level algorithm and Psuedocode~\ref{appendix:code:decomp} for low-level implementation}. 
In practice, \eqref{eq:monotonicity} may not hold for every step due to noise in the embedding {and pixel} space, and we find that a simple
{
pre-set minimal time interval between each decomposed subgoal to ensure they aren't too close together and a
}
low pass filter procedure to first smoothen the embedding distances make the subgoal criterion~\eqref{eq:subgoal-criterion} effective; 
see Appendix.~\ref{appnedix:sec:method} for more details.

\vspace{-0.5em}
\RestyleAlgo{ruled}
\begin{algorithm}[htbp!]
\SetKwInput{kwInit}{Init}
\caption{\ourmethod}\label{algo:decomp}
\kwInit{frozen visual encoder $\phi$, $\tau = \{o_0, \cdots, o_T\}$}
\kwInit{set of subgoals $\tau_{goal}$ = \{\}, $t = T$}
  \While{$t$ not small enough}{
    $\tau_{goal} = \tau_{goal} \cup \{o_{t}\}$ \\
    Find $o_{t-n-1}$ from Eq.~\ref{eq:subgoal-criterion} \\
    $ t = t - n - 1$ \\
  }
\end{algorithm}

{
In Fig.~\ref{fig:sim-real-wild-decomp}, we visualize \acronym's iterative decomposition process, showcasing the embedding distance curves (left) and the resulting decomposed subgoals (right). The featured videos are sourced from FrankaKitchen for the simulation task, \texttt{Fold-cloth} for the real-robot task, and a real-world sequence that includes: opening a drawer, picking the charger, plugging it in, activating the power strip, and partially closing the drawer. The black dotted lines indicate the timesteps where \acronym identifies a monotonic break, taking into account the minimal interval mentioned earlier. More visualizations can be found in Appendix~\ref{appendix:sec:qualitative-decomp}.

\noindent \textbf{Computational Efficiency.} We highlight that our entire algorithm does not require any additional neural network training or forward computations on top of the one forward pass required to encode all observations for policy learning. 
{
We further provide the decomposition and inference runtime for our \acronym implementation in Appendix.~\ref{appendix:method:runtime} The results indicate that \acronym introduces negligible costs ($<$ 0.1 sec. in total) when compared to standard policy learning methods.
}

\begin{figure*}[htbp!]
    \centering
    \includegraphics[width=0.95\linewidth]{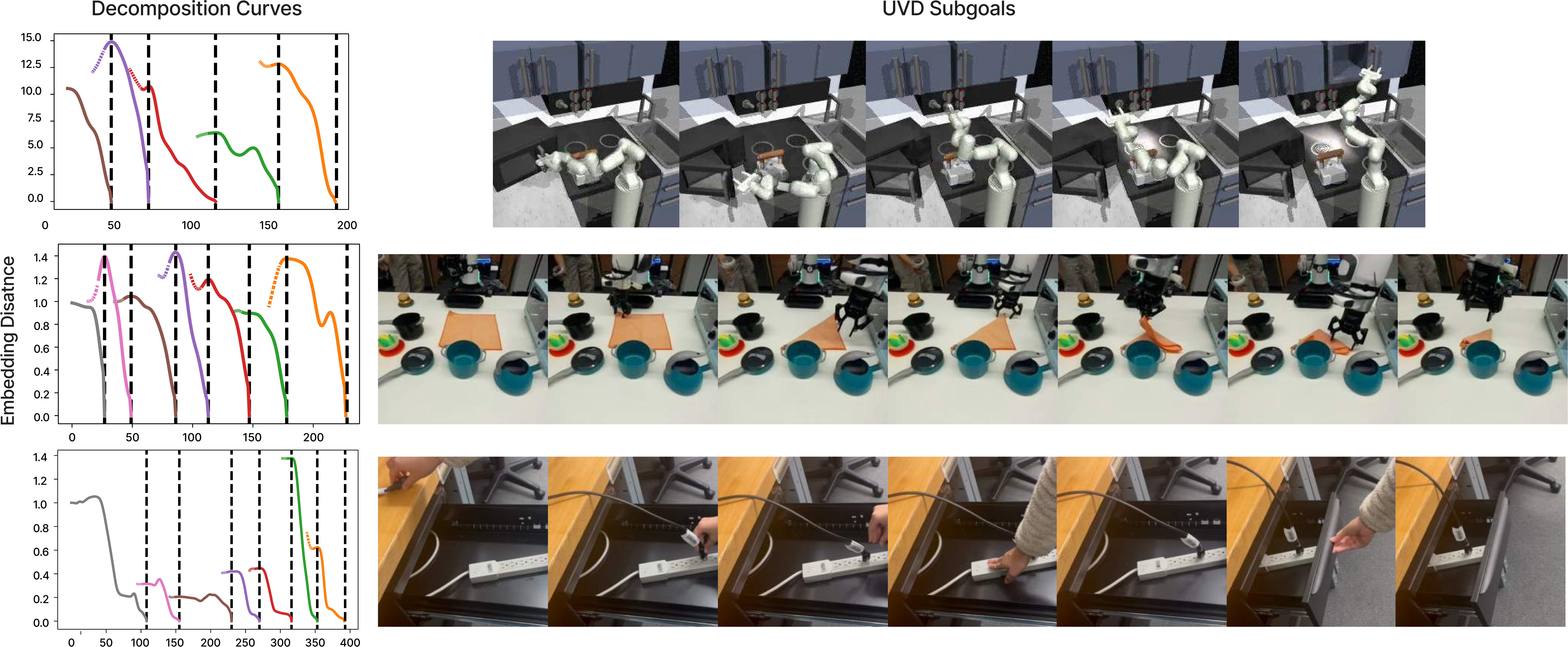}
    \caption{\textbf{Illustrations of \acronym's iterative decomposition, showcasing embedding distance curves (left) and decomposed subgoals (right).} The videos come from FrankaKitchen for the simulation task, \texttt{Fold-cloth} for the real-robot task, and a real-world video depicting the sequence: opening a drawer, picking the charger, plugging it in, activating the power strip, and partially closing the drawer.}
    \label{fig:sim-real-wild-decomp}
\end{figure*}

\subsection{\acronym-Guided Policy Learning}\label{subsec:method-learning}
Now, we discuss several ways \acronym-discovered subgoals can be used to supplement policy learning. 

\noindent \textbf{Goal Relabeling.} As \acronym is performed on a trajectory basis, we can relabel all observations in a trajectory with the closest subgoals that appear later in time. In particular, for an action-labeled trajectory $\tau=(o_0,a_0,...,o_T,a_T)$ and \acronym-discovered subgoals $\tau_{goal} = (g_0,...,g_m)$, we have that $\texttt{Label}(o_t)=g_k$ where $g_k$ is the first subgoal occurring after time $t$. This procedure leads to an augmented, goal-relabeled trajectory $\tau_{aug} = \{(o_0,a_0,g_0), ..., (o_T, a_T, g_m)\}$. Now, as all transitions are goal-conditioned, we can learn policies using any goal-conditioned imitation learning algorithm; for simplicity, we use goal-conditioned behavior cloning (GCBC)~\citep{ding2019goal, ghosh2019learning}.

\noindent \textbf{Reward Shaping.} The above goal relabeling strategy applies to the imitation learning (IL) setting. Collecting the demonstrations needed for IL is, however, expensive. Instead, a reinforcement learning paradigm is feasible with much fewer demonstrations and comes with other ancillary benefits such as learned error recovery. This raises the question of how \acronym-subgoals might be used with an RL paradigm. In particular, how can \acronym help overcome the exploration challenge in long-horizon RL? Given that \acronym selects subgoals so that the embedding distances in-between any two consecutive subgoals exhibit monotone trends, we define the \textit{UVD reward} to be the goal-embedding distance \textit{difference} computed using \acronym goals:
\begin{equation} 
\label{eq:embedding-reward}
R(o_t,o_{t+1};\phi, g_i) := d_{\phi}(o_{t};g_i)-d_{\phi}(o_{t+1};g_i) \ .
\end{equation}
where $g_i\in\tau_{goal}$, and $g_i$ will be switched to $g_{i+1}$ automatically during training when $d_{\phi}(o_{t+1};g_i)$ is small enough. 
More details can be found at Appendix.~\ref{appendix:subsec:rl}.
This choice of reward encourages making consistent progress towards the goal and has been found in prior work~\citep{wijmans2020pointnav,deitke2020robothor,weihs2021rearrangement,ma2022vip} to be particularly effective when deployed with suitable visual representations.

\subsection{\acronym Goal Inference}\label{subsec:method-goal-inference}
When deploying our trained subgoal-conditioned policies at inference time, we must determine what subgoals to instruct the policy to follow at each observation step. We study two simple strategies that work well in practice; we describe the high-level approaches here, and include more 
details on Appendix.~\ref{appendix:subsec:infer}.

\noindent \textbf{Nearest Neighbor.} First, when there is only one fixed sequence of subtasks to be learned (\ie, \indom), we employ a simple nearest neighbor goal selection strategy. That is, for a new observation, we compute the observation in the training set that has the closest embedding (judged by $d_\phi$) and use its associated sub-goal. This can be interpreted as a \textit{non-parameteric} high-level policy that outputs observation-conditioned goal for the low level policy, $\pi(\phi(o),\phi(g))$.

\noindent \textbf{Goal Relaying.} 
When performing \outdom or multi-task \indom evaluation, the agent must complete a user-instructed task. In these settings, the above nearest neighbor approach may no longer apply as the subgoals seen in training may not be valid for the current, potentially unseen, task. 
Instead, we propose to relay the currently instructed goals based on embedding distance. Specifically, given a sequence of instructed subgoals $g=(g_0,...,g_m)$, the policy will condition on the first remaining subgoal until the embedding distance between the current observation and the subgoal is below a certain threshold, at which point the policy will be conditioned on the next subgoal in the sequence.

\section{Experiments}

We study the following research questions:
\begin{enumerate}
    \item Does \acronym enable compositional generalization in multi-stage and multi-task imitation learning?
    \item Can \acronym subgoals enable reward-shaping for long-horizon reinforcement learning?
    \item Can \acronym be deployed on real-robot tasks?
\end{enumerate}

\begin{table*}[t]

\centering
\begin{tabular}{r|r|cccc}
\toprule
\textbf{Representation} & \textbf{Method} & \textbf{\indom success} & \textbf{\indom completion} & \textbf{\outdom success} & \textbf{\outdom completion} \\ \midrule 

\multirow{2}{*}{VIP (ResNet50)~\cite{ma2022vip}} 
    & GCBC & 0.736 (0.011)  & 0.898 (0.006) & 0.035 (0.014)  & 0.236 (0.057) \\
    & GCBC $+$ Ours & 0.737 (0.012) & 0.903 (0.009) & \bestscore{0.188 (0.024)} & \bestscore{0.566 (0.020)} \\
\midrule 

\multirow{2}{*}{R3M (ResNet50)~\cite{nair2022r3m}} 
    & GCBC & 0.742 (0.026) & 0.856 (0.006)  & 0.014 (0.007) & 0.223 (0.029) \\
    & GCBC $+$ Ours & 0.738 (0.024) & \bestscore{0.879 (0.000)}  & \bestscore{0.084 (0.045)} & \bestscore{0.427 (0.002)}  \\
\midrule 

\multirow{2}{*}{LIV (ResNet50)~\cite{ma2023liv}} 
    & GCBC & 0.608 (0.068) & 0.816 (0.046) & 0.008 (0.008) & 0.116 (0.082) \\
    & GCBC $+$ Ours & \bestscore{0.649 (0.013)} & \bestscore{0.868 (0.007)} & \bestscore{0.066 (0.025)} & \bestscore{0.496 (0.033)} \\
\midrule 

\multirow{2}{*}{CLIP (ResNet50)~\cite{khosla2020supervised}} 
    & GCBC & 0.391 (0.017)  & 0.692 (0.008) & 0.005 (0.001) & 0.119 (0.017) \\
    & GCBC $+$ Ours & 0.394 (0.036) & 0.701 (0.012) & \bestscore{0.073 (0.003)} & \bestscore{0.403 (0.01)} \\
\midrule

\multirow{2}{*}{DINO-v2 (ViT-large)~\cite{oquab2023dinov2}} 
    & GCBC & 0.329 (0.025) & 0.654 (0.019) &  0.012 (0.01) & 0.261 (0.213) \\
    & GCBC $+$ Ours & 0.322 (0.053) &  0.669 (0.037)   & \bestscore{0.055 (0.025)} & \bestscore{0.446 (0.034)} \\
\midrule 
\midrule
\multirow{2}{*}{VIP (ResNet50)~\cite{ma2022vip}} 
    & GCBC-GPT & 0.702 (0.029) & 0.841 (0.02) & 0.039 (0.027) & 0.302 (0.028) \\
    & GCBC-GPT $+$ Ours & 0.708 (0.056) & \bestscore{0.897 (0.024)} & \bestscore{0.213 (0.054)} & \bestscore{0.600 (0.038)}  \\

\bottomrule 
\end{tabular}
\caption{{\textbf{\indom and \outdom IL Results on FrankaKitchen.} We report the mean and standard deviation of success rate (full-stage completion) and the percentage of the completion (out of 4 stages), evaluated over diverse existing pretrained visual representations trained by GCBC with three seeds. Highlighted scores represent improvements in \outdom evaluations and \indom results with gains exceeding 0.01.}}
\vspace{-1em}
\label{table:ablation_encoder}
\end{table*}

\subsection{Simulation Experiments}\label{subsec:exp-sim}

\noindent \textbf{FrankaKitchen Environment.}
We use the FrankaKitchen Environment~\cite{gupta2019relay} for simulation experiments. In the environment, a Franka robot with a 9-DoF torque-controlled action space can interact with seven objects: a microwave, a kettle, two stove burners, a light switch, a hinge cabinet, and a sliding cabinet. We refined the dataset from \cite{gupta2019relay} to include only successful trajectories, yielding a total of 513 episodes gathered from humans using VR headsets. For each episode, four out of the seven objects are manipulated in an arbitrary sequence, leading to 24 unique completion orders; see Fig.~\ref{fig:fk_partition}.

\begin{figure}[t]
    \centering
    \includegraphics[width=1.0\linewidth]{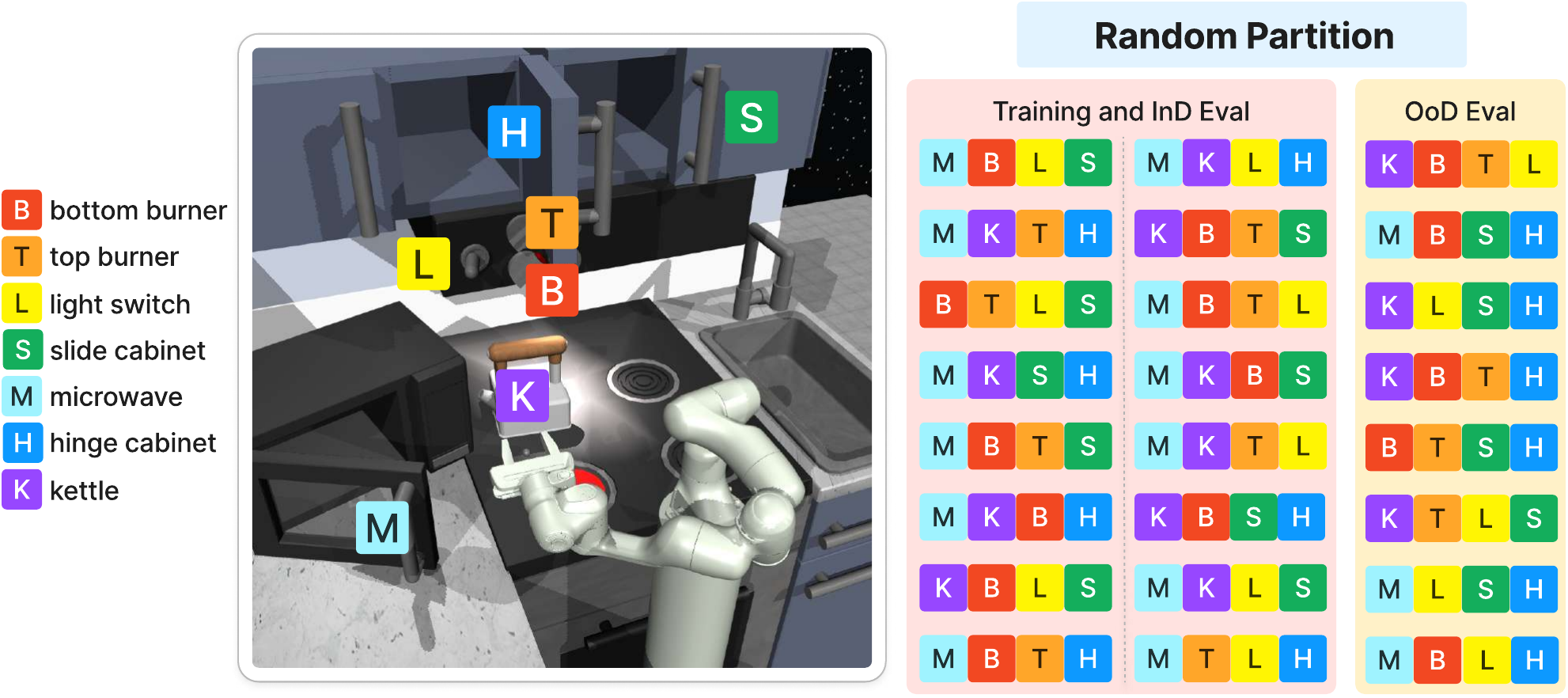}
    \caption{{\textbf{Frank Kitchen environment and an example of random training-evaluation partition.} In each demonstration episode, 4 out of 7 objects are manipulated in an arbitrary order. We show an example of 16 completion orders for training and \indom evaluation chosen randomly, while the rest of 8 are for \outdom generalizations. }}
    \label{fig:fk_partition}
    \vspace{-1em}
\end{figure}

\noindent \textbf{Visual and Policy Backbones.} 
As \acronym is designed to utilize pre-trained visual representations that capture visual task progress, we adopt R3M~\cite{nair2022r3m}, VIP~\cite{ma2022vip}, and LIV~\cite{ma2023liv}, three Resnet50-based representations trained with temporal objectives on video data; in particular, VIP and LIV are trained to explicitly encode smooth temporal task progress in their embedding distances. We also consider general vision models trained on static image datasets such as CLIP (ResNet50)~\cite{radford2021learning} and DINO-v2 (ViT-large)~\cite{oquab2023dinov2} to assess the importance of training on temporal data. As our goal is to study the merit of the pretrained representations, as in prior works~\citep{nair2022r3m, ma2022vip}, we keep the policy architecture simple and employ a multi-layer perception (MLP) as the policy architecture; 
More details in Appendix.~\ref{appendix:sec:arch}.

\noindent \textbf{Baselines.} 
We compare with goal-conditioned behavior cloning (GCBC) baselines to demonstrate the value of \acronym. Fixing a choice of visual representation, the only difference of GCBC to ours is the how the goals are labeled at training time. For each observation, GCBC labels the final frame in the same trajectory as its goal. 

\noindent \textbf{IL Evaluation Protocol.} 
Our training and evaluation design for FrankaKitchen is structured as follows: we train on $n$ combinations of object sequences with \indom evaluation, reserving the remaining $24-n$ task sequences for evaluation of unseen \outdom scenarios. We use $n=16$ by default unless otherwise mentioned. For a fair comparison, we utilize the same 3 random seen-unseen partitions, generated by 3 unique pre-defined seeds, for every set of runs. 

To evaluate policy performance, we consider both the success rate on the overall task (\textbf{success}) as well as the number of subtasks accomplished 
(\textbf{completion}). The success criterion for each subtask is determined by the simulation ground-truth state;
this is used solely during evaluations and is not provided to the agent during training. Results are presented in Table~\ref{table:ablation_encoder}.

\noindent \textbf{Results.} Remarkably, by using \acronym, \textbf{\textit{all}} pre-trained visual representation show significant improvement in \outdom sequential generalization, despite their varying \indom performances. VIP and LIV, the two representations explicitly trained to learn monotone embedding distances, demonstrate higher \textit{comparative} gains compared to the other representations, despite similar or even lower performances when the representations are not used to decompose subgoals (i.e., GCBC-MLP); this validates our hypothesis that representations capturing visual task progress information are more suited for off-the-shelf subgoal discovery. \newline 

\begin{figure}[htbp!]
    \centering
    \includegraphics[width=\linewidth]{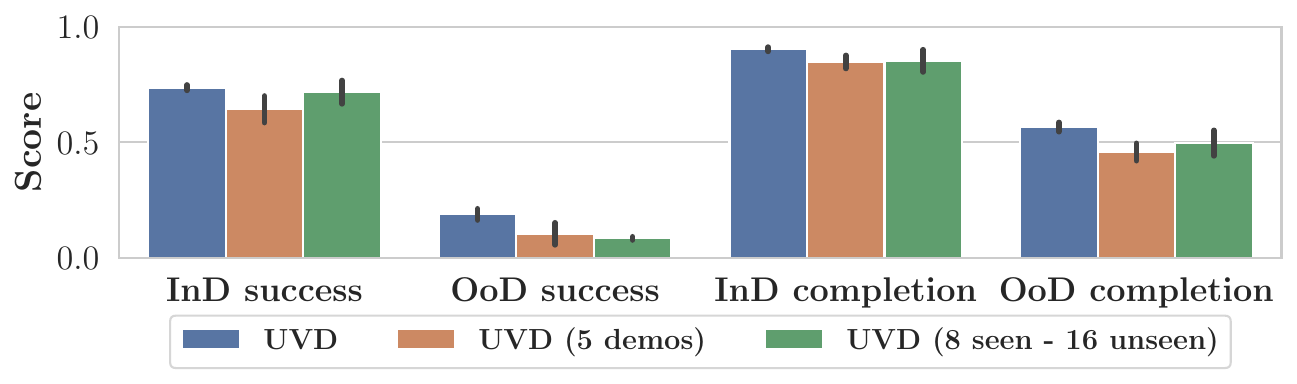}
    \caption{{\textbf{Ablations on dataset size and composition.}}}
    \label{fig:fk_heuristic}
    \vspace{-1em}
\end{figure}

\noindent \textbf{Ablations.} We present several ablations studying whether \acronym remains effective when varying training settings. As VIP stands out as the most promising candidate for \acronym-based imitation learning, we perform all ablations using VIP as the backbone representation. First, we ablate the MLP policy architecture with a GPT-like causal transformer policy~\citep{radford2019language}. As shown in the last row of Table~\ref{table:ablation_encoder}, this more powerful, history-aware, policy is insufficient to achieve the same level of generalization; \acronym again provides sizable generalization improvement.

Beyond policy architecture, we also study the effect of dataset size and diversity. To this end, we consider (1) reducing the training dataset size to $5$ demonstrations per training task, and (2) reducing the number of training tasks to $8$ but keeping the full number of demonstrations per task. Both \indom and \outdom performance remains similar, confirming that \acronym enables \outdom generalization that is robust to the varying sizes and diversity of the training data.

\begin{figure}[htbp!]
    \centering
    \includegraphics[width=1\linewidth]{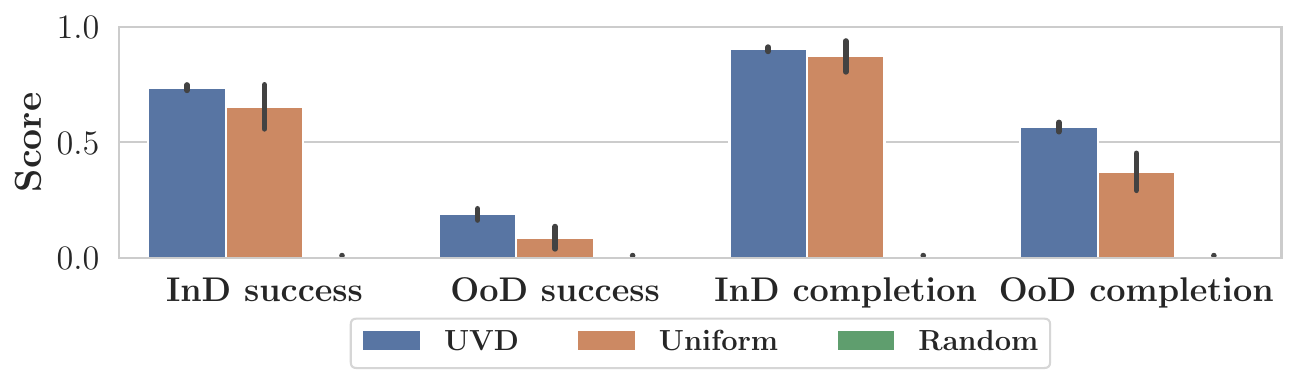}
    \vspace{-1em}
    \caption{{\textbf{Comparison with heuristic goal-labeling methods.}}}
    \label{fig:fk_heuristic}
\end{figure}

Finally, we study whether \textit{UVD} is necessary to achieve strong \outdom generalization and investigate alternative ways of generating subgoals. We consider \textbf{Uniform} and \textbf{Random};
\textbf{Uniform} randomly selects a frame within a fixed size window after the observation; this strategy has been employed in many prior works~\citep{gupta2019relay, lynch2020learning}. \textbf{Random} randomly selects 3 to 5 frames within the demonstration as subgoal frames.
As shown in Fig.~\ref{fig:fk_heuristic}, the alternatives uniformly hurt performance on all settings and metrics. This is to be expected as these alternatives introduce redundant and less semantically meaningful subgoals; as a result, they may perform comparably \indom, but their \outdom generalization suffers.

\begin{figure*}[t]
  \centering
    \includegraphics[width=1\textwidth]{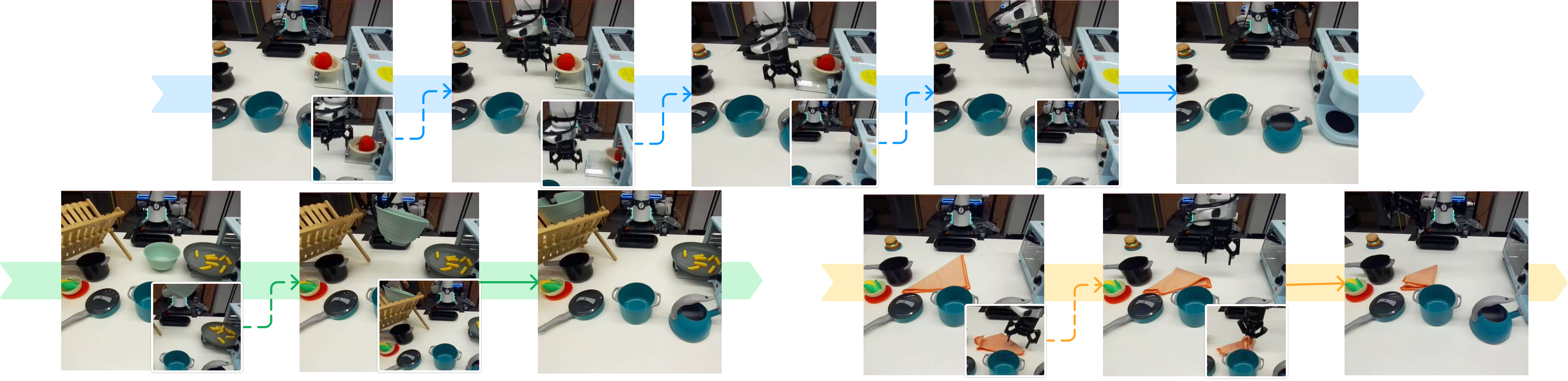}   %
  \caption{{\textbf{Example Sub-Sampled Rollouts on Real-World OOD Tasks.} The initial frame in each sequence is a representative \outdom initial observation. The inset image in each frame is the conditioned \acronym-discovered goal for that frame.}}
  \label{fig:real_robot_ood}
\end{figure*}

\begin{table}[t]
\vspace{-1em}
\centering
\begin{tabular}{@{}l@{\hspace{1em}}c@{\hspace{1em}}c}
\toprule
 \textbf{Method} & \textbf{Success} & \textbf{Completion} \\
 \midrule 
     GCRL-VIP & 0.0 / 0.0 & 0.09 / 0.25  \\
     GCRL-VIP $+$ Ours & \bestscore{0.65 / 1.0} &  \bestscore{0.75 / 1.0} \\
\midrule 
    GCRL-R3M & 0.0 / 0.0 & 0.09 / 0.25  \\
    GCRL-R3M $+$ Ours & \bestscore{0.649 / 1.0} & \bestscore{0.82 / 1.0}  \\
\bottomrule 
\end{tabular}
\caption{{\textbf{RL results on FrankaKitchen}. Full-stage success rate and the percentage of full-stage completion are reported in the format of (average performance with 3 random seeds) / (max performance).}}
\vspace{-1em}
\label{table:rl-results}
\end{table}

\noindent \textbf{\acronym-Guided Reinforcement Learning.} We investigate whether \acronym can also enhance reinforcement learning by providing goal-based shaped rewards for subtasks~\eqref{eq:embedding-reward}. Recall that in this setting, only a single video demonstration (without action labels) is given to the agent to specify the learning task. Within the FrankaKitchen environment, we examine a specific task sequence: \texttt{open microwave}, \texttt{move kettle}, \texttt{toggle light switch}, and \texttt{slide cabinet}. We select VIP and R3M as candidate representations as they performed best for IL \indom evaluations. We consider a goal-conditioned RL baseline, which constructs goal-based rewards by uniformly using the last demonstration frame as goal in~\eqref{eq:embedding-reward}.
We use PPO~\citep{schulman2017proximal} as the RL algorithm and report the average and the max success rate and percentage of completion over 3 random seeds in Table~\ref{table:rl-results}; 
see Appendix.~\ref{appendix:subsec:rl} for more details.

We see baselines fail to make non-trivial task progress with either visual backbone, confirming that goal-based rewards with respect to a distant final goal are not well-shaped to guide exploration. In contrast, \acronym-rewards consistently accelerate RL training and achieve high overall success on the task, validating \acronym's utility in not only task generalization but also in task learning.

\subsection{Real-World Experiments}\label{subsec:real-exp}

\begin{figure}[t]
    \centering
    \includegraphics[width=\linewidth]{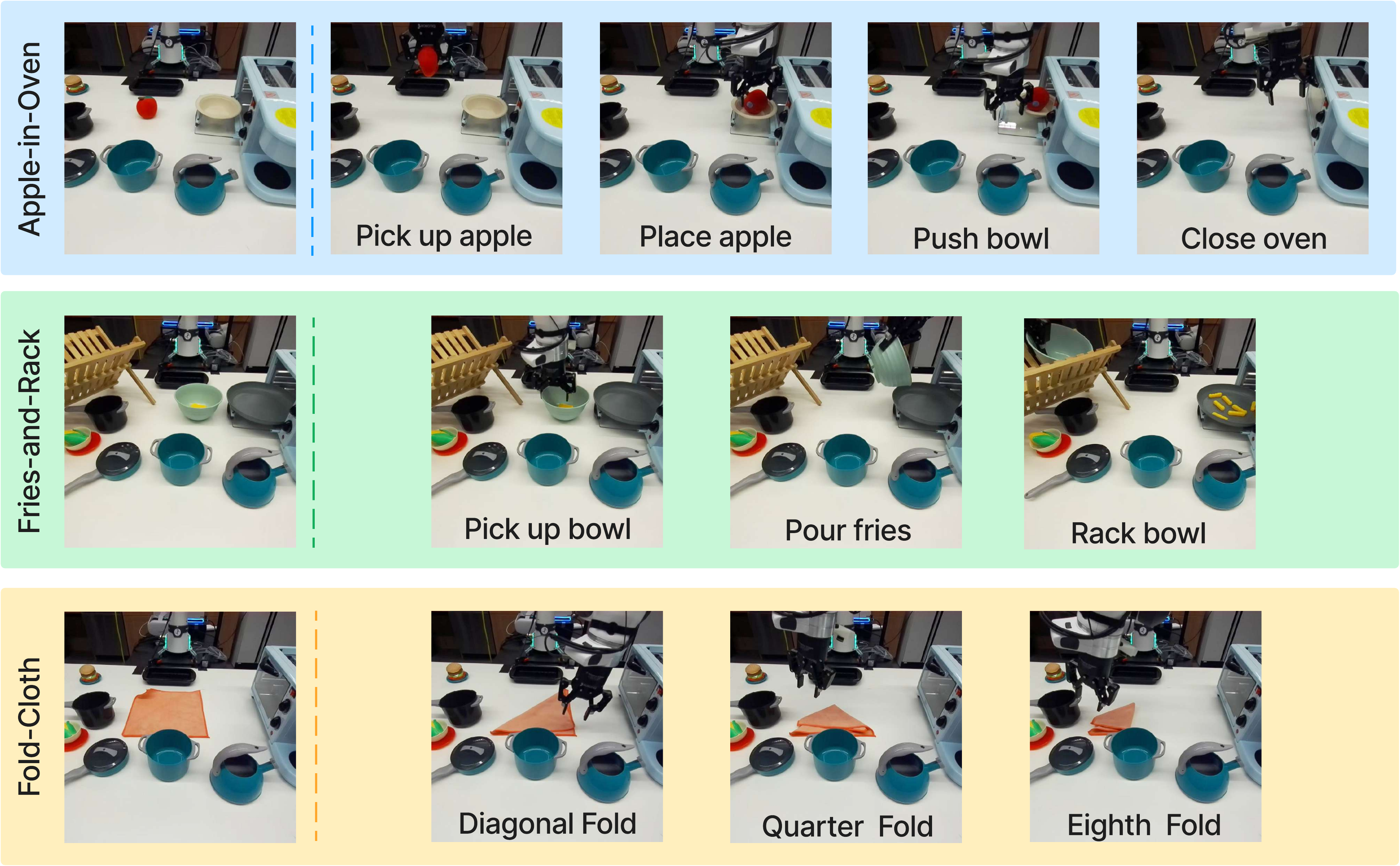}
    \caption{{\textbf{Real-World Tasks.} The first picture in each row depicts a representative initial observation, and the following frames are the distinct subtasks.}}
    \label{fig:real_tasks}
    \vspace{-1em}
\end{figure}

\begin{table}[t]
\vspace{1em}
\centering
\resizebox{1.0\columnwidth}{!}{
\begin{tabular}{c|r|cccc}
\toprule
\textbf{Task} & \textbf{Method} & \textbf{\indom S.} & \textbf{\indom C.} & \textbf{\outdom S.} & \textbf{\outdom C.} \\ \midrule 

\multirow{2}{*}{\texttt{Apple-in-Oven}} 
    & GCBC &  0.50 & 0.438  &  0.0 &  0.500 \\
    & GCBC $+$ Ours & 0.60 & \bestscore{0.750}  & \bestscore{0.25} &  \bestscore{0.625}\\
\midrule 

\multirow{2}{*}{\texttt{Fries-and-Rack}} 
    & GCBC & 0.30 & 0.567  & 0.0 & 0.0\\
  & GCBC $+$ Ours & 0.35 & \bestscore{0.750} & \bestscore{0.25} &  \bestscore{0.500}\\
\midrule 

\multirow{2}{*}{\texttt{Fold-Cloth}} 
    & GCBC &  0.05& 0.100 &  0.0&  0.0\\
    & GCBC $+$ Ours & 0.15 & \bestscore{0.483} & \bestscore{0.15}&  \bestscore{0.425}\\

\bottomrule 
\end{tabular}}
\caption{{\textbf{\indom and \outdom Results on Real-World Tasks.}} S-success, C-completion.}
\vspace{-1em}
\label{table:real-world}
\end{table}

We introduce 3 real-world multi-stage tasks on a real Franka robot. These tasks contain daily household manipulation skills, such as picking, pouring, folding, and manipulating articulated objects. See Fig.~\ref{fig:real_tasks} for a detailed breakdown of the subtasks in each task. For each task, we have collected about 100 demonstrations via teleoperation; for each trajectory, the positions of relevant objects in the scene are randomized within a fixed distribution. The policies are learned via GCBC with MLP architecture as in simulation; 
see Appendix.~\ref{appendix:fig:realsetup} for more real-robot details.

\noindent \textbf{\outdom Evaluation.} On our real-world tasks, the subtasks are sequentially dependent and cannot be performed in arbitrary orders. To test compositional generalization, we evaluate whether the policies can \textit{skip} intermediate tasks when their effects in the environment are already achieved. For example, on the \texttt{Fries-and-Rack} task, we evaluate on initial states in which the fries are already placed on the plate. In this case, a policy trained with semantically meaningful subgoals should be able to directly proceed from picking up the bowl to racking the bowl. This is because the post-condition of pouring the fries is semantically identical to the pre-condition of racking the bowl -- both have the bowl picked up mid air and the fries on the plate. Similarly, on the \texttt{Apple-in-Oven} task, we test generalization by having the apple directly placed on the plastic plate, and on the \texttt{Fold-Cloth} task, we have the cloth folded diagonally already; see Figure~\ref{fig:real_robot_ood} for an illustration of these \outdom initial observations. While these \outdom tasks are shorter than the training tasks, the exact sequences are still unseen during training and they contain unseen initial state configurations. As before, we test these \outdom as well as \indom task sequences; for each task sequence, we evaluate on 20 rollouts using the same set of object configurations for every compared method.

Results are presented in Table~\ref{table:real-world}. As shown, on all tasks, UVD methods can solve \outdom tasks whereas the baseline completely fails, despite their comparable performance on \indom tasks. These results corroborate our findings in simulation and make a strong case for the effectiveness of UVD's subgoals and its applicability to challenging real-world tasks. 
{
In addition to introducing \outdom for initial states, we also show in Appendix~\ref{appendix:extend_exp-real} that our plug-and-play can recover from unexpected human interference in intermediate states. In contrast, GCBC struggles significantly in these situations. This underscores the ability of \acronym's subgoals to guide the agent effectively in challenging \outdom situations.
}

In Figure~\ref{fig:real_robot_ood}, we visualize \acronym policy rollouts by displaying sub-sampled frames and their conditioned subgoals (the inset frame) on the \outdom tasks. In all cases, \acronym retrieves meaningful subgoals from the training set and the policy can successfully match the depicted semantic subtask.

\section{Limitation and Future Works}

While \acronym offers the advantage of not necessitating any task-specific knowledge or training, its efficacy is well-demonstrated across both simulated and real-robot environments. However, as we only validate on fully observable manipulation tasks, direct application to navigation tasks, especially those embodied tasks involving partial observations, may not yield intuitive or explainable subgoals (even though representations are pretrained with temporal objective using egocentric datasets~\cite{nair2022r3m, ma2022vip, ma2023liv}). 

Looking ahead, we are eager to broaden the applications of \acronym, diving deeper into its capabilities within egocentric scenarios, and even the key-frame extraction for video understanding and dense video caption tasks. On another front, while task graphs are widely used in Reset-Free RL~\cite{gupta2021reset, xu2023dexterous}, acquiring milestones as subgoals is resource-intensive and lacks scalability. By integrating our off-the-shelf \acronym subgoals into the task-graph, we are interested in seeing agents that, with minimal resets, can adeptly handle a wide range of tasks across various sequences and horizons in the wild.

\section{Conclusion}

We have presented \ourmethod, an off-the-shelf task decomposition method for long-horizon visual manipulation tasks using pre-trained visual representations. \acronym does not require any task-specific knowledge or training and effectively produces semantically meaning subgoals across both simulated and real-robot environments. UVD-discovered subgoals enable effective reward shaping for solving challenging multi-stage tasks using RL, and policies trained with IL exhibit significantly superior compositional generalization at test time.

\section*{Acknowledgments}
{
We would like to thank Aniruddha Kembhavi, Ranjay Krishna, Yunfan Jiang, Oscar Michel, Jiasen Lu, and many other colleagues for their insightful feedback and discussions.
}
This work is funded in part by NSF CAREER Award 2239301, NSF Award CCF-1917852, and ARO Award W911NF-20-1-0080. The U.S. Government is authorized to reproduce and distribute reprints for Government purposes notwithstanding any copyright notation herein.

\newpage

\bibliographystyle{IEEEtranS}
\bibliography{root}

\clearpage

\appendices
\setcounter{table}{0}
\renewcommand{\thetable}{\thesection.\arabic{table}}
\setcounter{figure}{0}
\renewcommand{\thefigure}{\thesection.\arabic{figure}}

\setcounter{AlgoLine}{0}
\renewcommand{\theAlgoLine}{\thesection.\arabic{AlgoLine}}

\setcounter{lstlisting}{0}
\renewcommand{\thelstlisting}{\thesection.\arabic{lstlisting}}

\section*{Appendix}

\section{Details on \acronym}\label{appnedix:sec:method}

\subsection{Decomposition}

As we show in Algo.~\ref{algo:decomp}, UVD preprocesses the demonstration or user-provided raw video offline. We further provide low-level pseudocode in Python of \acronym in Pseudocode.~\ref{appendix:code:decomp}. In practice, when identifying the temporally nearest observation where the monotonicity condition is not met, as per Eq.~\ref{eq:subgoal-criterion}, it is equivalent to locating the most recent local maximum of the embedding distance curves. This is because the distance curve is anticipated to show an almost monotonically decreasing trend towards the final frame in each recursive iteration, as shown in Eq.~\ref{eq:monotonicity}. Due to the small-scale noise in pixel and high-dimensional feature space, we apply Nadaraya-Watson Kernel Regression~\cite{takeda2007kernel} to first smooth the embedding curves before calculating the embedding distances. We benchmark our UVD implementation runtime in Appendix.~\ref{appnedix:sec:method}\ref{appendix:method:runtime}, which shows a negligible time span, even when handling high-resolution videos with substantial duration in the wild.

\begin{lstlisting}[language=Python,label={appendix:code:decomp},caption={\textbf{\acronym implementation in Python}}]
from scipy.signal import argrelextrema

def UVD(
    embeddings: np.ndarray | torch.Tensor, 
    smooth_fn: Callable,
    min_interval: int,
) -> list[int]:
    # last frame as the last subgoal
    cur_goal_idx = -1 
    # saving (reversed) subgoal indices (timesteps)
    goal_indices = [cur_goal_idx]
    cur_emb = embeddings.copy() # L, d
    while cur_goal_idx > min_interval:
        # smoothed embedding distance curve (L,)
        d = norm(cur_emb - cur_emb[-1], axis=-1)
        d = smooth_fn(d)
        # monotonicity breaks (e.g. maxima)
        extremas = argrelextrema(d, np.greater)[0]
        extremas = [
            e for e in extremas 
            if cur_goal_idx - e > min_interval
        ]
        if extremas:
            # update subgoal by Eq.(3)
            cur_goal_idx = extremas[-1] - 1
            goal_indices.append(cur_goal_idx)
            cur_emb = embeddings[:cur_goal_idx + 1]
        else:
            break
    return embeddings[
        goal_indices[::-1]  # chronological
    ] 
\end{lstlisting}

In all of our simulation and real-world experiments, we use $\texttt{min\_interval}=20$, and Radial Basis Function (RBF) with the bandwidth of $0.08$ for Kernel Regression, to eliminate most of the visual and motion noise in the video. We provide UVD decomposition qualitative results in Appendix.~\ref{appendix:sec:qualitative-decomp}.

\subsection{Inference}\label{appendix:subsec:infer}

We now elucidate the specifics of applying UVD subgoals in a multi-task setting during inference. Remember that given a video demonstration represented as $\tau = (o_0, \cdots, o_T)$ and UVD-identified subgoals $\tau_{goal} = (g_0, \cdots, g_m)$, we can extract an augmented trajectory labeled with goals, represented as $\tau_{aug} = {(o_a, a_0, g_0), \cdots, {o_T, a_T, g_m}}$. This is useful for goal-conditioned policy training, as discussed in Sec.~\ref{subsec:method-learning}.

For inference, we can similarly produce an augmented offline trajectory without the necessity of ground-truth actions, i.e., $\tau_{aug,infer} = \{(o_0, g_0), \cdots, (o_T, g_m)\}$. In the online rollout, after resetting the environment to $o_0$, the agent continuously predicts and enacts actions conditioned on subgoal $g_0$ using the trained policy. This continues until the embedding distance between the current observation and the subgoal surpasses a pre-set positive threshold $\epsilon$ at a specific timestep $i$, \ie $d_\phi(o_i; g_0) < \epsilon$, where $\phi$ is the same frozen visual backbone used in decomposition and training. Following this, the subgoal will be seamlessly transitioned to the next, continuing until success or failure is achieved. 

In practice, the straightforward goal-relaying inference method might face accumulative errors during multiple subgoal transitions, especially due to noise from online rollouts. However, when an agent is guided explicitly by tasks depicted in a video, incorporating the duration dedicated to each subgoal can help reduce this vulnerability. To clarify, once we've aligned subgoals with observations from the video, we also draw a connection between the timesteps of observations and their corresponding subgoals. We denote the \textit{subgoal budget} for subgoal $g_i = o_t$ as $\mathcal{B}_{g_i} := n + 1$ where $g_{i-1} = o_{t-n-1}$ based on Eq.~\ref{eq:subgoal-criterion}. 
Building on this, we propose a secondary criterion for switching subgoals: verify if the relative steps completing the current stage are in the neighborhood of the subgoal budget. This measure ensures timely transitions: it avoids prematurely switching before completing a sub-stage or delaying the transition despite accomplishing the sub-stage in the environment. To sum up, given an ongoing observation $o_t$ and subgoal $g_i$ at timestep $t$, and considering the preceding subgoal $g_{i-1}$ at timesteps $t - h$, the subgoal will transition to $g_{i+1}$ if
\begin{equation}
d_\phi(o_t; g_i) < \epsilon \quad \text{and} \quad |h - \mathcal{B}_{g_i}| < \delta    
\end{equation}
We use $\epsilon=0.2$ and $\delta=2$ steps for all of our experiments, except in baseline tests that are conditioned solely on final goals.

\subsection{Feature Continuity}
We then visualize 3D t-SNE in feature space from different frozen visual backbones in Fig.~\ref{appendix:fig:tsne}. We include VIP~\cite{ma2022vip}, R3M~\cite{nair2022r3m}, CLIP~\cite{radford2021learning}, and ResNet~\cite{he2016deep} trained for ImageNet-1k classification~\cite{deng2009imagenet}. As shown in Fig.~\ref{appendix:fig:tsne}, representations pretrained with temporal objectives, \eg VIP~\cite{ma2022vip} and R3M~\cite{nair2022r3m}, provide more smooth, continuous, and monotone clusters than other vision foundation models. In practice, those representations with more smooth and continuous embedding curves provide better \acronym decomposition as well as better performance in downstream control.

\begin{figure*}[htbp!]
    \centering
    \includegraphics[width=1.0\linewidth]{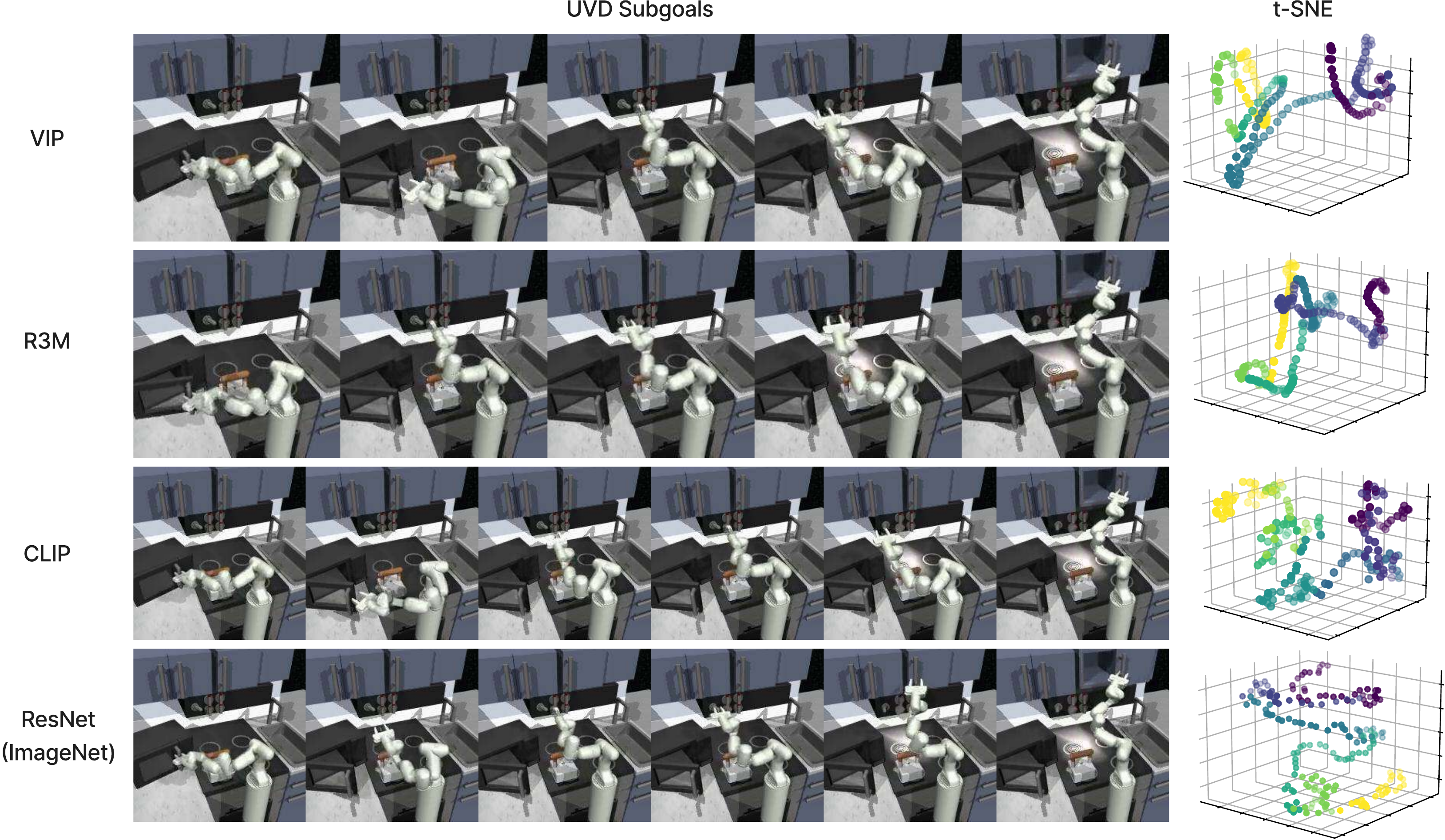}
    \caption{\textbf{\acronym subgoals and 3D t-SNE visualizations of different frozen visual embeddings.} t-SNE colors are labeled by \acronym subgoals.
    Representations pretrained with temporal objectives like VIP~\cite{ma2022vip} and R3M~\cite{nair2022r3m} provide more smooth, continuous, and monotone clusters in feature space than others, whereas the ResNet trained for supervised classification on ImageNet-1k~\cite{deng2009imagenet} provide the most sparse embeddings.}
    \label{appendix:fig:tsne}
    \vspace{-1em}
\end{figure*}

\subsection{Runtime of \acronym}\label{appendix:method:runtime}

Finally, We present the average runtime of \acronym for preprocessing and decomposing trajectories. We break the runtime by: 1) \textbf{load} from raw video file into an array; 2) \textbf{preprocess} the video array by the frozen visual encoder (including the tensor and device conversions); and 3) apply \textbf{\acronym} to the preprocessed embeddings. In addition to the FrankaKitchen dataset and real-world data used for our experiments, we assessed the decomposition capabilities on a 720P MOV video. The video has a frame rate of $30$ fps, contains $698$ frames (equating to a duration of $23.3$ seconds), and is decomposed into 16 subgoals. Visualizations can be found in Fig.~\ref{appendix:fig:decomp-in-the-wild-longest}. Runtimes were calculated based on an average of 513 episodic data from the FrankaKitchen dataset and 100 trials for the \texttt{in the wild} video, all processed on an RTX A6000 GPU.
Preprocessing is required only once offline before the policy training. As indicated in Tab.~\ref{appendix:tab:uvd_runtime}, \acronym operates in a negligible time span, even when handling high-resolution videos with substantial duration in the wild.

\begin{table}[htbp!]

\centering
\begin{tabular}{r|c|ccc}
\toprule

 & \textbf{\# frames}  & \textbf{Load} & \textbf{Preprocess} & \textbf{\acronym} \\
\midrule 
\texttt{FrankaKitchen} & 226.9 & 0.023 & 0.155 & 0.0023 \\
\texttt{In the wild} & 698 & 1.011 & 0.450 & 0.011 \\

\bottomrule 
\end{tabular}
\caption{\textbf{\acronym offline preprocess runtimes (in seconds).} }
\label{appendix:tab:uvd_runtime}
\end{table}

\section{Models}\label{appendix:sec:arch}

\subsection{Policies} 
To underscore that our method serves as an off-the-shelf method that is applicable to different policies, we ablate with a Multilayer Perceptron (MLP) based single-step policy and a GPT-like causal transformer policy. We summarize the MLP and GPT policies hyperparameters in Tab.~\ref{appendix:tab:MLP_param}~\ref{appendix:tab:GPT_param}. The MLP policy, akin to the designs in~\cite{nair2022r3m, ma2022vip} for downstream control tasks, employs a 3-layer MLP
with hidden sizes of [1024, 512, 256] to produce deterministic actions. 
This MLP ingests a combination of the frozen visual embeddings from step-wise RGB observations and goal images followed by a 1D BatchNorm, as well as the 9D proprioceptive data encoded through a single layer complemented by a LayerNorm. 

Our GPT policy removes the BatchNorm and replaces the MLP with the causal self-attention blocks consisting of 8 layers, 8 heads, and an embedding dimension of 768. We set an attention dropout rate of 0.1 and a context length of 10. 
The implementation is built upon~\cite{karpathy2023nanoGPT, touvron2023llama}.
We transition from the conventional LayerNorm to the Root Mean Square Layer Normalization (RMSNorm)~\cite{zhang2019rmsnorm} and enhance the transformer with rotary position embedding (RoPE)~\cite{su2021roformer}. Actions are predicted via a linear similar to~\cite{chen2021decision}. In practice, we generally found this recipe has more stable training and better performance than the original implementation from~\cite{karpathy2023nanoGPT, shafiullah2022behavior}. At inference time, we cache the keys and values of the self-attention at every step, ensuring that there's no bottleneck as the context length scales up. Nevertheless, in the FrankaKitchen tasks, we observed that a longer context length tends to overfit and performance drop. Therefore, we consistently use a context length of 10 for all experiments.

\begin{table}[htbp!]

\centering
\begin{tabular}{r|cccc}
\toprule
\footnotesize

\textbf{Policy} & \textbf{MLP} & \textbf{MLP + \acronym} & \textbf{GPT} & \textbf{GPT + \acronym} \\
\midrule 
Episodic Runtime & 6.03 & 6.17 & 7.43 & 7.50 \\

\bottomrule 
\end{tabular}
\caption{\textbf{Benchmark the inference runtime (in seconds).} Runtimes are averaged across 100 rollouts with one GPU process and episodic horizon 300.}
\label{appendix:tab:inference_runtime}
\end{table}

\section{Training Details}

\subsection{Imitation Learning}

We summarize the hyperparameters of imitation learning in Tab.~\ref{appendix:tab:il_param}.
In the simulation, we conduct an online evaluation every 100 epochs using 10 parallel environments for each GPU machine. We choose the best checkpoints based on the combined \indom and \outdom performance, averaged across all multi-task training scenarios. For benchmarking inference time, we employ a single GPU, comparing our approach with the GCBC baseline as shown in Tab.~\ref{appendix:tab:inference_runtime}. This further underscores that \acronym incurs negligible overhead throughout the preprocessing, training, and inference stages.

\begin{table*}[tbp!]
\centering
\begin{minipage}{.3\linewidth}
    \centering
    \begin{tabular}{@{}ccc@{}}
        \toprule
        \textbf{Hyperparameter/Value} & \textbf{MLP-Policy} & \textbf{GPT-Policy} \\ \midrule
        Optimizer & AdamW~\cite{loshchilov2017decoupled} & AdamW~\cite{loshchilov2017decoupled} \\
        Learning Rate & 3e-4 & 3e-4 \\
        Learning Rate Schedule & cos decay & cos decay \\
        Warmup Steps & 0 & 1000 \\
        Decay Steps & 150k & 200k \\
        Weight Decay & 0.01 & 0.1 \\
        Betas & [0.9, 0.999] & [0.9, 0.99] \\
        Max Gradient Norm & 1.0 & 1.0 \\
        Batch Size & 512 & 128 \\
        \bottomrule
    \end{tabular}
    \caption{\textbf{IL training hyperparameters}}
    \label{appendix:tab:il_param}
\end{minipage}
\hfill
\begin{minipage}{.2\linewidth}
    \centering
    \begin{tabular}{@{}cc@{}}
        \toprule
        \textbf{Hyperparameter} & \textbf{Value} \\ \midrule
        Hidden Dim. & [1024, 512, 256] \\
        Activation & ReLU \\
        Proprio. Hidden dim. & 512 \\
        Proprio. Activation & Tanh \\
        Visual Norm. & Batchnorm1d \\
        Proprio. Norm. & LayerNorm \\
        Action Activation & Tanh \\
        Trainable Parameters & 3.3M \\
        \bottomrule
    \end{tabular}
    \caption{\textbf{MLP policy hyperparameters}}
    \label{appendix:tab:MLP_param}
\end{minipage}
\hfill
\begin{minipage}{.3\linewidth}
    \centering
    \begin{tabular}{@{}cc@{}}
        \toprule
        \textbf{Hyperparameter} & \textbf{Value} \\ \midrule
        Context Length & 10 \\
        Embedding Dim. & 768 \\
        Layers & 8 \\
        Heads & 8 \\
        Embedding Dropout & 0.0 \\
        Attention Dropout & 0.1 \\
        Normalization & RMSNorm~\cite{zhang2019rmsnorm} \\
        Action Activation & Tanh \\
        Trainable Parameters & 58.6M \\
        \bottomrule
    \end{tabular}
    \caption{\textbf{GPT policy hyperparameters}}
    \label{appendix:tab:GPT_param}
\end{minipage}
\end{table*}

\subsection{Reinforcement Learning}\label{appendix:subsec:rl}

All RL experiments are trained using the Proximal Policy Optimization (PPO)~\cite{schulman2017proximal} RL algorithm implemented within the AllenAct~\cite{weihs2020allenact} RL training framework. We summarize the hyperparameters of reinforcement learning in Tab.~\ref{appendix:tab:rl_param}. 

In our RL setting, the configurations for both training and inference remain consistent. This is analogous to the inference for IL as detailed in Appendix.~\ref{appnedix:sec:method}\ref{appendix:subsec:infer}. Specifically, the task is also specified by an unlabeled video trajectory $\tau$. Given the initial observation $o_0$ and UVD subgoal $g_0\in\tau_{goal}$, the agent continuously predicts and executes actions conditioned on subgoal $g_0$ using the online policy with frozen visual encoder $\phi$, until the condition $d_\phi(o_t;g_0) < \epsilon$ satisfied for some timestep $t$ and positive threshold $\epsilon$. 

As shown in Eq.~\ref{eq:embedding-reward}, we provide progressive rewards defined as goal-embedding distance \textit{difference} using \acronym subgoals. Recognizing that the distance between consecutive subgoals can vary, we employ the normalized distance function: $\bar{d_\phi}(o_t;g_i) := d_\phi(o_t;g_i) / d_\phi(g_{i-1}; g_i)$. This ensures that $\bar{d_\phi}(o_{t-h};g_i) \approx 1$ for some timestep $t-h$ that the subgoal was transitioned from $g_{i-1}$ to $g_i$. Additionally, we provide modest discrete rewards for encouraging (chronically) subgoal transitions, and larger terminal rewards for the full completion of task sub-stages, which is equivalent as the embedding distance between the observation and the final subgoal becomes sufficiently small. To sum up, at timestep $t$, the agent is receiving a weighted reward
\begin{equation}\label{eq:literal_rl_rewards}
\begin{aligned}
    R_t &= \alpha\cdot\left(\bar{d_\phi} (o_{t-1};g_i) - d_\phi(o_t;g_i) \right) \\
    &+ \beta \cdot \mathbf{1}_{\bar{d_\phi}(o_t;g_i) < \epsilon} \\
    &+ \gamma \cdot \mathbf{1}_{\bar{d_\phi}(o_t;g_m) < \epsilon} 
\end{aligned}
\end{equation}
based on the RGB observations $o_t, o_{t-1}$, corresponding \acronym subgoal $g_i\in\tau_{goal}$, and final subgoal $g_m\in\tau_{goal}$. While similar reward formulations appear in works such as~\cite{ng1999policy,lee2021generalizable,li2022phasic,ma2022vip,zhang2023learning}, we are the first in delivering optimally monotonic implicit rewards unsupervisedly by UVD, derived directly from RGB features. In our experiments, we use $\alpha=5, \beta=3, \gamma=6, \epsilon=0.2$, and confine the first term within the range $[-\alpha, \alpha]$ in case edge cases in feature space. For the final-goal-conditioned RL baseline, it is equivalent as $g_i = g_m = o_{T}\in \tau = \{o_0, \cdots, o_T\}$ and $\beta=0$ in Eq.~\ref{eq:literal_rl_rewards}.

Tab.\ref{table:rl-results} illustrates that the simple incorporation of UVD-rewards greatly enhances performance. We also showcase a comparison of evaluation rewards between GCRL and GCRL augmented with our UVD rewards. This is done using the R3M~\cite{nair2022r3m} and VIP~\cite{ma2022vip} backbones, as seen in Fig.~\ref{appendix:fig:rl-reward}. This highlights the capability of our UVD to offer more streamlined progressive rewards. This capability is pivotal for the agent to adeptly manage the challenging, multi-stage tasks presented in FrankaKitchen. To the best of our knowledge, ours is the first work to achieve such a high success rate in the FrankaKitchen task without human reward engineering and additional training. Notably, our RL agent, trained with the optimally monotonic \acronym-reward, can complete 4 sequential tasks in as few as \textbf{90 steps} --- a stark contrast to the over 200 steps observed in human-teleoperated demonstrations. This further illustrates the \acronym-reward's potential to encourage agents to accomplish multi-stage goals more efficiently. The videos of rollouts can be found on our website.

\begin{table}[htbp!]
\centering
\begin{tabular}{@{}cc@{}}
    \toprule
    \textbf{Hyperparameter} & \textbf{Value} \\ \midrule
    GPU instances & 8x RTX A6000 \\
    Environments per GPU & 8 \\
    Optimizer & AdamW~\cite{loshchilov2017decoupled} \\
    Learning rate & 3e-4 \\
    Learning rate schedule & linear decay \\
    Max gradient norm & 0.5 \\
    Discount factor $\gamma$ & 0.99 \\
    GAE $\tau$ & 0.95 \\
    Value loss coefficient & 0.5 \\
    Normalized advantages & True \\
    Entropy coefficient & 0.001 \\ 
    Rollout length & 200 \\
    PPO epochs & 10 \\
    Number of mini-batches & 1 \\
    PPO Clip & 0.1  \\
    \bottomrule
\end{tabular}
\caption{\textbf{RL hyperparameters.}}
\label{appendix:tab:rl_param}
\end{table}

\begin{figure}[htbp!]
    \centering
    \vspace{-1em}
    \includegraphics[width=0.9\linewidth]{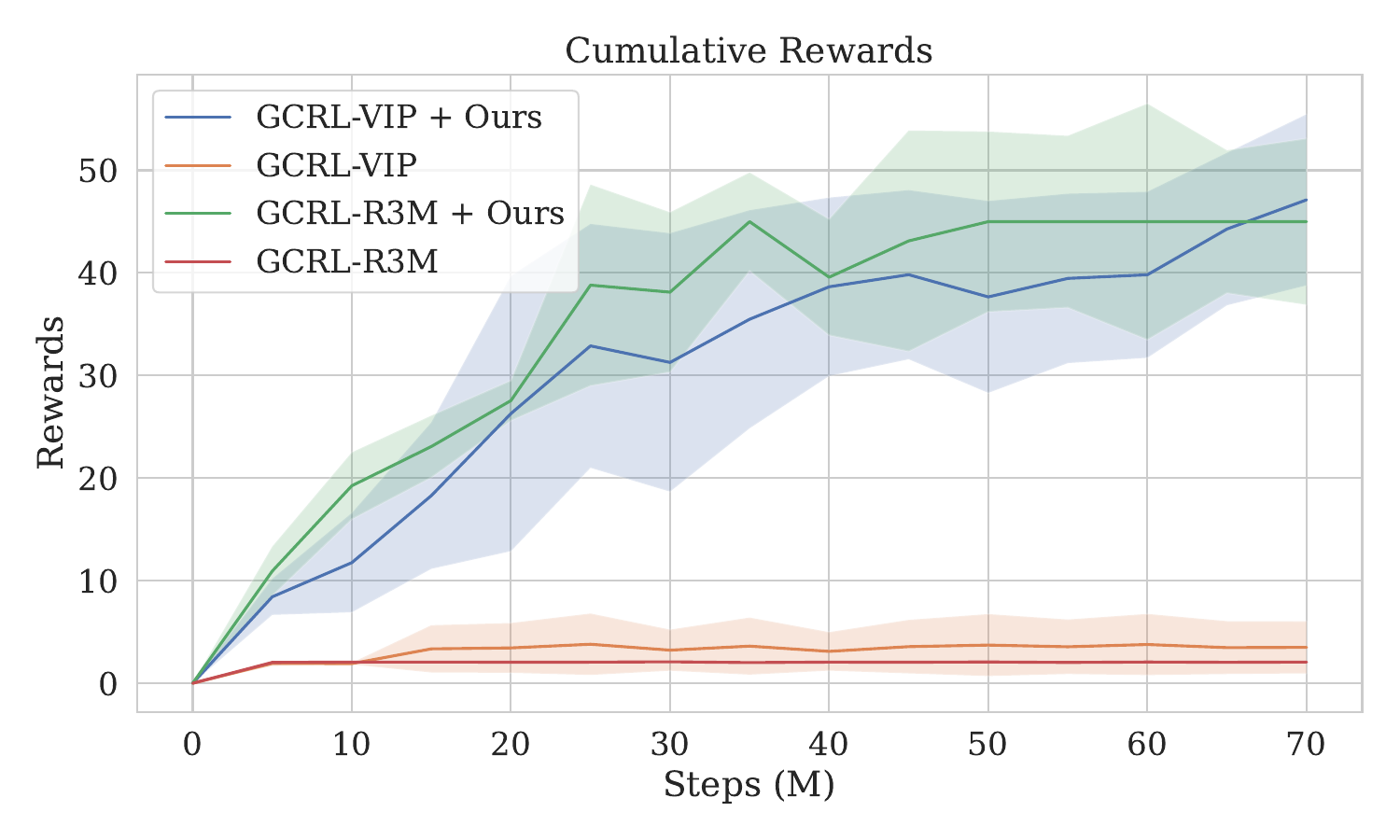}
    \caption{\textbf{RL evaluation cumulative rewards.} Note that different visual backbones may not be comparable due to different representation spaces, but we show significant progressive signals in comparison with the final-goal-conditioned baseline.}
    \label{appendix:fig:rl-reward}
    \vspace{-1em}
\end{figure}

Failures in the GCRL baselines predominantly stem from the agent's tendency to get trapped in local minima, usually when it achieves a task sub-stage that results in the most significant visual changes in the feature space, \eg sliding the cabinet in this case. Since it causes the most noticeable shifts both in pixel and feature representations, the baseline agent often fixates on this subtask alone, making no further progress. Conversely, when employing UVD subgoals and rewards, we have observed a marked difference during training. The agent incrementally learns to navigate the entire task, approaching it stage by stage. These stages are not isolated, as they share pertinent information. For example, UVD breaks down task sequences into phases characterized by nearly monotonic motions. These can be categorized as the ``hand reaching" or ``object interaction" phases.  This shared knowledge framework means that once the agent masters the initial ``hand reaching" phase, subsequent similar hand-reaching motions become more intuitive to learn and execute. Nevertheless, we do occasionally observe instances where applying \acronym results in failure. In these cases, the agent often oscillates its gripper back and forth, seemingly hacking the reward shaping, which in turn leads to an irreversible state. We speculate that incorporating supervised human intervention~\cite{singh2022ask4help} or unsupervised near-irreversible detection~\cite{zhang2023learning}, could address this issue and further enhance performance.
\section{Extended Experiments and Ablations}

\begin{table*}[htbp!]

\centering
\begin{tabular}{r|r|cccc}
\toprule
\textbf{Representation} & \textbf{Method} & \textbf{\indom success} & \textbf{\indom completion} & \textbf{\outdom success} & \textbf{\outdom completion} \\ \midrule

\multirow{2}{*}{VIP (5 demos)} 
    & GCBC-GPT & 0.409 (0.102) & 0.702 (0.066) & 0.005 (0.005) & 0.285 (0.024) \\
    & GCBC-GPT $+$ Ours & \bestscore{0.419 (0.027)} & \bestscore{0.763 (0.016)} & \bestscore{0.13 (0.033)}  & \bestscore{0.533 (0.026)} \\
\midrule 

\multirow{2}{*}{VIP (5 demos)} 
    & GCBC-MLP & \bestscore{0.668 (0.024)} & 0.82 (0.035)  & 0.016 (0.016) & 0.208 (0.006) \\
    & GCBC-MLP $+$ Ours & 0.643 (0.058) & \bestscore{0.848 (0.028)} & \bestscore{0.104 (0.048)}  & \bestscore{0.458 (0.038)} \\
\midrule 

\multirow{2}{*}{VIP (8 seen - 16 unseen)} 
    & GCBC-MLP & 0.724 (0.057) & 0.851 (0.036) &  0.001 (0.001) & 0.102 (0.020) \\
    & GCBC-MLP $+$ Ours & 0.717 (0.051) & 0.853 (0.048) & \bestscore{0.084 (0.007)}  & \bestscore{ 0.497 (0.055)} \\
\midrule 

\multirow{2}{*}{VIP (8 seen - 16 unseen)} 
    & GCBC-GPT & 0.602 (0.114) & 0.554 (0.48)  & 0.003 (0.003)  & 0.143 (0.124) \\
    & GCBC-GPT $+$ Ours & 0.587 (0.049) & 0.558 (0.483)  &  \bestscore{0.037 (0.040)}  & \bestscore{0.307 (0.268)} \\
\midrule

\bottomrule 
\end{tabular}
\caption{\textbf{Ablations on dataset size and compositions, with comparisons with GCBC baselines and \acronym}}
\vspace{-1em}
\label{appendix:table:ablation}
\end{table*}

\subsection{Simulation}

We present numerical results for ablations from Sec.~\ref{subsec:exp-sim} in Tab.~\ref{appendix:table:ablation}, with extended comparison with GCBC baselines. Without surprise, our method consistently outperforms baselines in compositional generalization settings, when varying the dataset size to $5$ demonstrations for each FrankaKitchen task, or adjusting the seen-unseen partitions, which doubles the count of unseen sequences while halving the number of seen ones.

In the FrankaKitchen demonstrations, there are 24 task sequences encompassing 4 subtasks each, translating to 4 ordered object interactions. We further compare GCBC$+$\acronym with GCBC conditioned on the final frames of each human-defined subtask in FrankaKitchen, utilizing both MLP and GPT policies. The main distinction between UVD-decomposed subgoals and the subgoals for each subtask is that UVD furnishes milestones emphasizing monotone motions, while subtasks yield subgoals with oracle semantic meanings.
Both methodologies employ the identical inference approach as described in Appendix~\ref{appendix:subsec:infer}, and share the same tasks partitions from three distinct seeds. The average successes and subtask completion rates are shown in Fig.~\ref{appendix:fig:fk_subtask_comp}. Surprisingly,  even when subgoals from sub-tasks offer ground-truth semantic meaning (and, notably, share identical conditioned subgoal frames across different task sequences), only GPT policy allows for performance surpassing our method in the \outdom setting. In the \indom setting, however, using subgoals from pre-defined sub-tasks leads to a substantial performance drop for MLP policies. This decline might be due to potential confusion (\eg mis-manipulated top and bottom burners) arising from similar subgoals when jointly training multiple multi-stage tasks when utilizing a lightweight, single-step MLP policy.

\begin{figure}[t]
    \centering
    \includegraphics[width=1.0\linewidth]{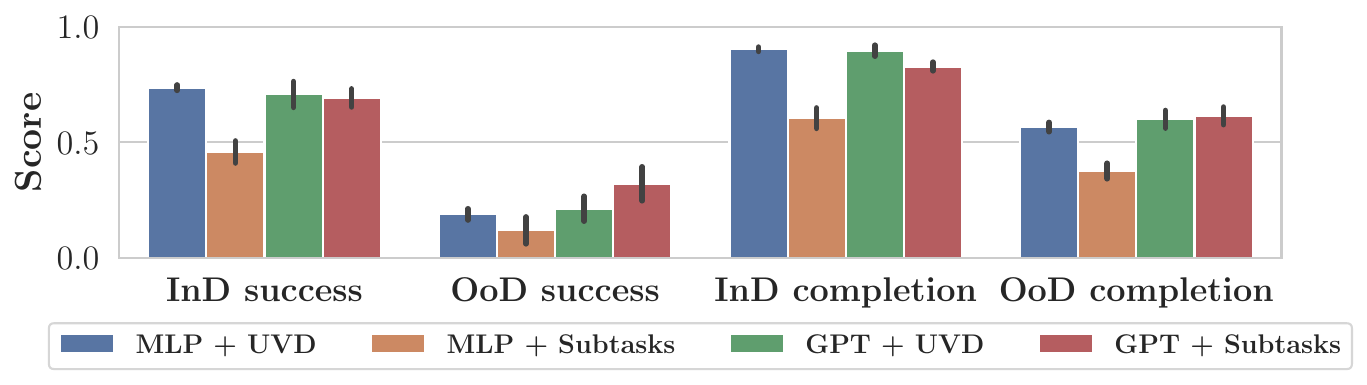}
    \caption{\textbf{Comparison with the decomposition from human pre-defined}}
    \label{appendix:fig:fk_subtask_comp}
    \vspace{-1em}
\end{figure}

\subsection{Real-Robot}\label{appendix:extend_exp-real}

In the \outdom settings, even though we have not achieved exceptionally high success rates for all of the tasks, we still managed a completion rate of around or above 50\%. This outcome can be attributed to the fact that policies trained using our method consistently exhibit the right intent. Instead of overfitting to the \indom settings, they tend to successfully complete intermediate steps and occasionally face challenges only at later stages.
In contrast, the GCBC baseline always overfits the \indom initial state. For example, in \texttt{Fold-Cloth} generalization experiments, the baseline still goes to the corner that is already folded. For more rollout visualizations, please refer to the videos available on our website.

We further extend our \outdom setting, which aimed for unseen initial states in Sec.~\ref{subsec:real-exp}, also encompass more diverse intermediate states. Our objective during deployment is to ensure the agent remains resilient to tasks, even in the presence of human interference. In $\texttt{Apple-in-Oven}$ and $\texttt{Fries-and-Rack}$ tasks, we introduce two more \outdom scenarios. In the first, we revert the scene by placing the apple back to its original position, challenging the agent to recover from this change. In the second, we manually circumvent an intermediate step. For instance, after the robot has grasped the bowl of fries, we manually transfer all the fries to the plate. This alteration means the agent should subsequently place the bowl directly on the rack without the need for pouring.

\begin{table}[htbp!]
\centering
\begin{tabular}{@{}ccc@{}}
    \toprule
    \textbf{Method} & \textbf{\texttt{Apple-in-Oven}} & \textbf{\texttt{Fries-and-Rack}}\\ \midrule
    GCBC &  0.0 & 0.2\\
    GCBC + Ours & \bestscore{0.5} & \bestscore{0.9}\\
    \bottomrule
\end{tabular}
\caption{\textbf{Success rate over 10 rollouts with human interference.}}
\label{appendix:tab:human_interference}
\end{table}
\section{Real-World Robot Experiment Details}

The robot learning environment is illustrated in Fig.~\ref{appendix:fig:realsetup}. We use a 7-DoF Franka robot with a continuous joint-control action space. A Zed 2 camera is positioned on the table's right edge, and only its RGB image stream—excluding depth information—is employed for data collection and policy learning. Another Zed mini camera is affixed to the robot's wrist. For the \texttt{Apple-in-Oven} task, we utilize the right view from both cameras, while for the \texttt{Fries-and-Rack} and \texttt{Fold-Cloth} tasks, we rely on their left views.

\begin{figure}[t]
    \centering
    \includegraphics[width=1.0\linewidth]{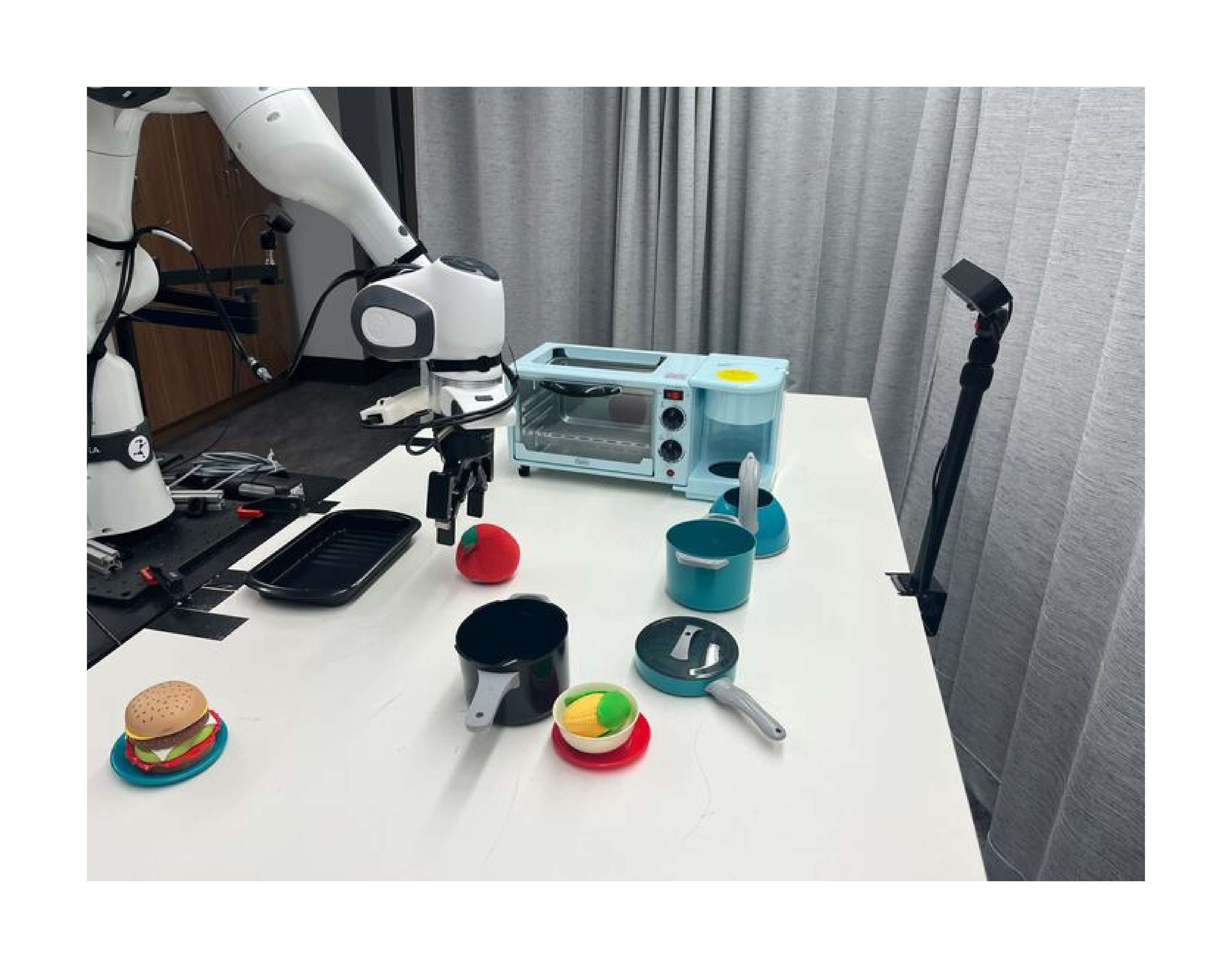}
    \caption{\textbf{Real robot experiments setup.}}
    \label{appendix:fig:realsetup}
    \vspace{-1em}
\end{figure}

\begin{figure*}[t]
    \centering
    \includegraphics[width=1.0\linewidth]{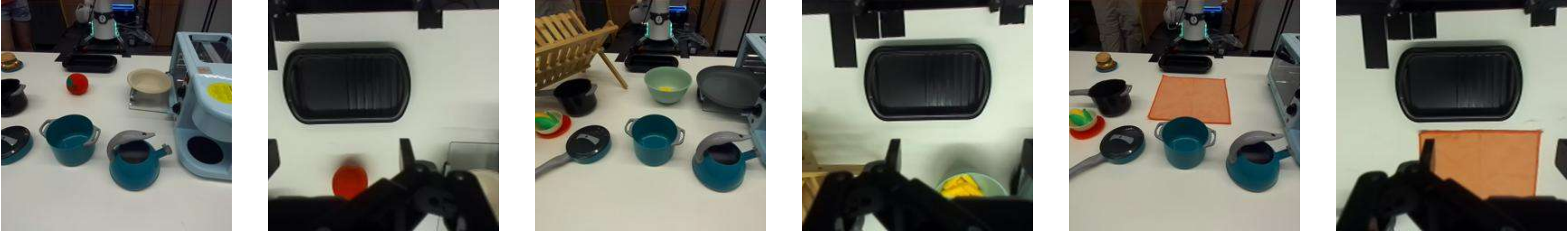}
    \caption{\textbf{Raw observations from two different cameras for three tasks.} Three (No.1, 3, 5 from the left) are from the side-view Zed2 camera and the others (No.2, 4, 6) are from Zed mini on the wrist.}
    \label{appendix:fig:raw_observations}
    \vspace{-1em}
\end{figure*}

\subsection{Task Descriptions}

\renewcommand\theadalign{c}
\renewcommand\theadgape{\Gape[2pt]}
\renewcommand\cellgape{\Gape[2pt]}

\begin{table}[htbp!]

\centering
\begin{tabular}{r|c|c}
\toprule
 & \textbf{EL} & \textbf{\# demos} \\
 \midrule
\texttt{Apple-in-Oven} & 197.5 & 105\\
\texttt{Fries-and-Rack} &  170.1 & 110 \\
\texttt{Fold-Cloth} & 246.8 & 105\\
\bottomrule
 
\end{tabular}
\caption{\textbf{Real Tasks average episode length (EL) and the number of demos (\# demos).} }
\label{appendix:tab:real_task}

\end{table}

We specify the average episode lengths and the number of demonstrations we used for experiments for each task in Tab.~\ref{appendix:tab:real_task}.
The criteria for successful task completion are as follows:
\begin{itemize}
    \item \texttt{Apple-in-Oven}: pick up the apple on the table; place the apple in the bowl without tipping it over; push the bowl into the oven; close the oven door.
    \item \texttt{Fries-and-Rack}: pour fries onto the plate, ensure at least half of them are on the plate; place the bowl on the rack without causing any collisions.
    \item \texttt{Fold-Cloth}: grasp the corner of the cloth and fold the cloth in the directions shown in the demonstrated video multiple times.
\end{itemize}
During evaluation, we assess the successful completion of each sub-stage over 20 rollouts to determine the overall success and completion rates.
To evaluate our policy's compositional generalization abilities, we introduce unseen initial states for each task. While the success criteria remain consistent with prior assessments, the initial step has been pre-completed by humans. The extended \outdom setting, which includes human interference in intermediate states, is detailed in Appendix~\ref{appendix:extend_exp-real}.

\subsection{Training and evaluation details}

\noindent \textbf{UVD in training:} As with our simulation experiments, we preprocess all the demos using UVD.
The behavior cloning policy further incorporates the view from the wrist camera besides the view from the side camera and decomposed subgoals during training.
Operating under velocity control, our robot's action space encompasses a 6-DoF joint velocity and a singular dimension of the gripper action (open or close). Consequently, the policy produces 7D continuous actions.
The robot control frequency is set as 15 Hz.

\noindent \textbf{UVD in evaluating:} In the training stage, we save the set of subgoals and corresponding observations over all demos in a task. During inference, every time the robot gets a pair of observations, we retrieve the subgoal that has the nearest observation ($l_2$ norm over observation embeddings) with the current one as the current subgoal. Then we concatenate the current observation with the retrieved subgoal together as input then get real-time joint velocity action.

Our method frequently results in the development of more robust policies, enabling recovery actions when initial attempts fail. For example, in the failure scenario of \texttt{Apple-in-Oven} and one of the successful cases in \texttt{Fold-Cloth} as showcased on our website, the policy opts for a reattempt, pushing the bowl further if not adequately placed inside the oven, or re-folding if the cloth's corner is not grasped properly. In contrast, such recovery behaviors are conspicuously absent in the GCBC baselines, further highlighting its propensity to overfit to \indom training setting.

\noindent \textbf{BC Model details:} We train our policy on a laptop with RTX 3080 GPU. For both the baseline policy and our method, we add proprioception to help learning and augment each training dataset by randomly cropping the input images. Since we have limited demonstrations in the real world, we only set MLP size to be [256,256]. Please refer to Tab.~\ref{appendix:tab:bc_param} for details.

\begin{table}[htbp!]
\centering
\begin{tabular}{@{}cc@{}}
    \toprule
    \textbf{Hyperparameter} & \textbf{Value} \\ \midrule
    GPU Instances & RTX 3080 Ti Laptop GPU \\
    MLP Architecture & [256, 256] \\
    Non-Linear Activation & ReLU \\
    Optimizer & 
    AdamW~\cite{loshchilov2017decoupled} \\
    Gradient Steps & 10k\\
    Batch Size & 64\\
    Learning Rate & 1e-3\\
    Proprioception & Yes \\
    Augmentation & Random crop \\
    \bottomrule
\end{tabular}
\caption{\textbf{Real robot BC hyperparameters.}}
\label{appendix:tab:bc_param}
\end{table}

\section{Qualitative Subgoal Decomposition Results}\label{appendix:sec:qualitative-decomp}

We show decomposition results with UVD on simulation videos in Fig.~\ref{appendix:fig:sim1} ~\ref{appendix:fig:sim2}, real robots videos in Fig.~\ref{appendix:fig:apple},~\ref{appendix:fig:fries},~\ref{appendix:fig:cloth}, and wild videos in Fig.~\ref{appendix:fig:closet},~\ref{appendix:fig:computer},~\ref{appendix:fig:drawer},~\ref{appendix:fig:wash},~\ref{appendix:fig:decomp-in-the-wild-longest}. From the subgoal decomposition results, we can have a clear overview of the key frames within a video. For instance, in \texttt{Fold-Cloth} video, UVD precisely catches the picking and placing key frames for three times.

Fig.~\ref{appendix:fig:sim1}~\ref{appendix:fig:sim2} are from our simluation experiments, Fig.~\ref{appendix:fig:apple},~\ref{appendix:fig:fries},~\ref{appendix:fig:cloth} are from real robot experiments. Fig.~\ref{appendix:fig:closet} video depicts a human opening a cabinet and rearranging items, while Fig.~\ref{appendix:fig:computer} video showcases unlocking a computer in an office. Furthermore, Fig.~\ref{appendix:fig:drawer} demonstrates the process of opening a drawer and charging a device, and Fig.~\ref{appendix:fig:wash} illustrates washing and then wiping hands in a bathroom. Lastly, Fig.~\ref{appendix:fig:decomp-in-the-wild-longest} presents activities shot in the kitchen with relatively longer duration. Based on the analysis of all video decomposition results, it is evident that UVD extends beyond robotic settings, proving to be highly effective in household scenarios captured in human videos.

\begin{figure*}[htbp]
    \centering
    \includegraphics[width=1.0\textwidth]{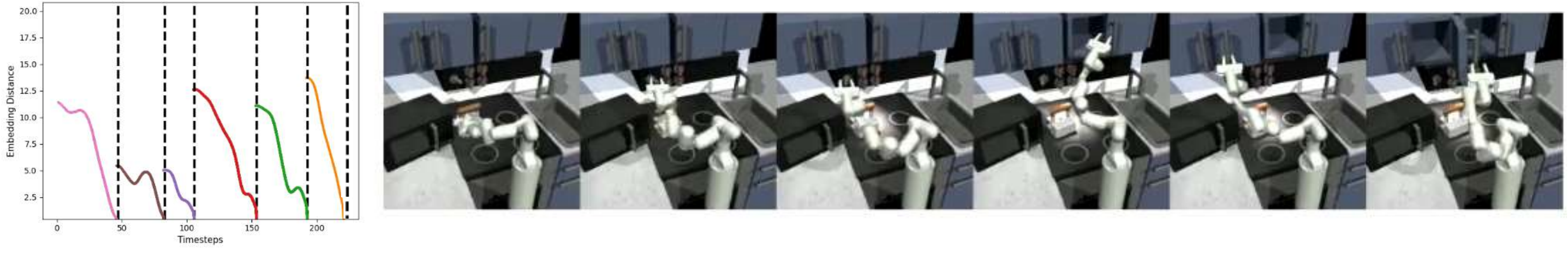}
    \label{fig:sub1}
  \caption{\textbf{Video sequence}: moving a kettle, turning on light switch, operating slide cabinet, operating hinge cabinet.}
  \label{appendix:fig:sim1}
\end{figure*}

\begin{figure*}
    \centering
    \includegraphics[width=0.75\textwidth]{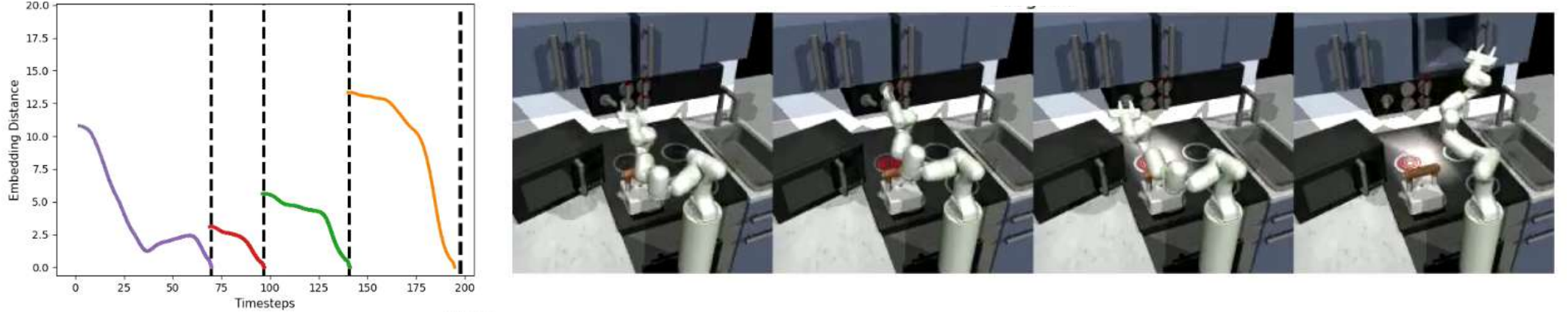}
    \caption{\textbf{Video sequence}: rotating bottom burner, rotating top burner, turning on light switch, operating slide cabinet.}
    \label{appendix:fig:sim2}
\end{figure*}

\begin{figure*}
    \centering
    \includegraphics[width=1.0\textwidth]{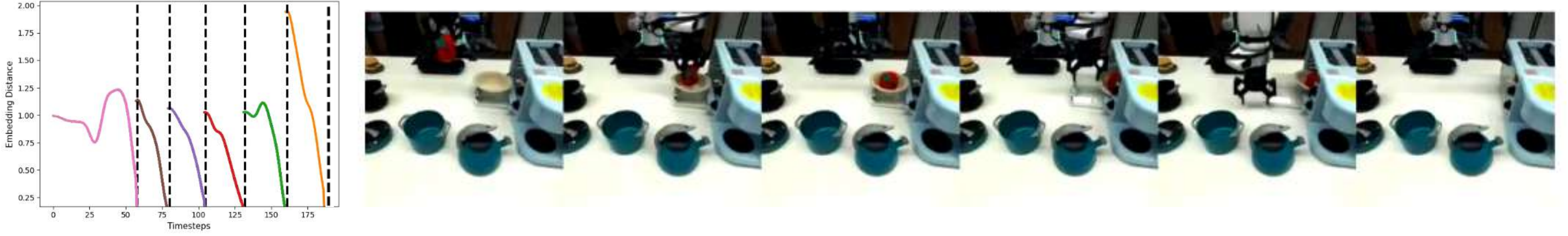}
    \caption{\textbf{Video sequence}: picking apple, placing apple in the bowl, pushing the bowl into the oven, closing the oven.}
    \label{appendix:fig:apple}
\end{figure*}

\begin{figure*}
    \centering    \includegraphics[width=0.85\textwidth]{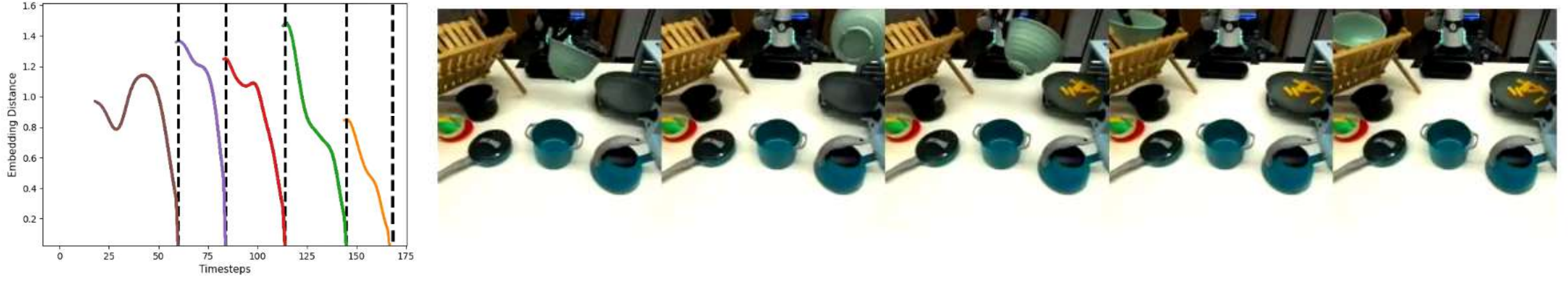}
    \caption{\textbf{Video sequence}: picking a bowl, pour fries out of the bowl, placing the bowl on the rack.}
    \label{appendix:fig:fries}
\end{figure*}

\begin{figure*}
    \centering
    \includegraphics[width=1.0\textwidth]{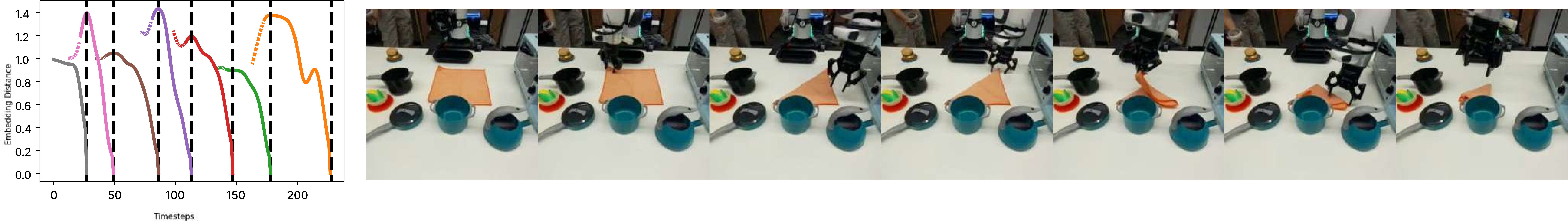}
    \caption{\textbf{Video sequence}: diagonal fold, quarter fold, eighth fold.}
    \label{appendix:fig:cloth}
\end{figure*}

\begin{figure*}
    \centering
    \includegraphics[width=0.8\textwidth]{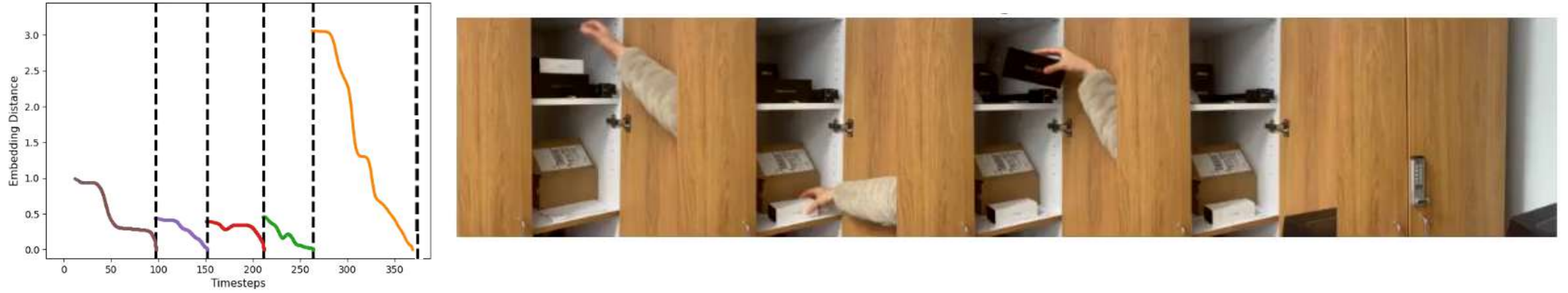}
    \caption{\textbf{Video sequence}: picking upper white box, placing white box, picking upper black box, closing the cabinet.}
    \label{appendix:fig:closet}
\end{figure*}

\begin{figure*}
    \centering
    \includegraphics[width=0.8\textwidth]{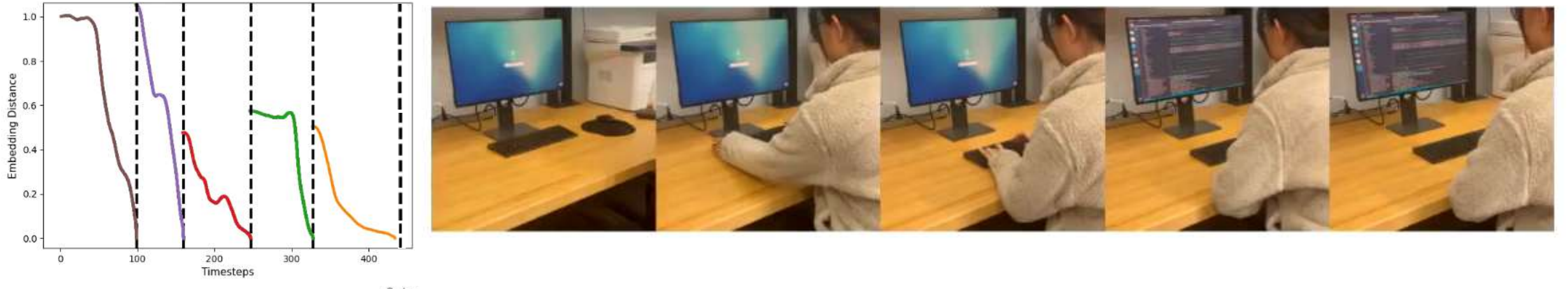}
    \caption{\textbf{Video sequence}: grabbing a chair, moving the keyboard towards human, typing password, unlocking a computer.}
    \label{appendix:fig:computer}
\end{figure*}

\begin{figure*}
    \centering
    \includegraphics[width=1.0\textwidth]{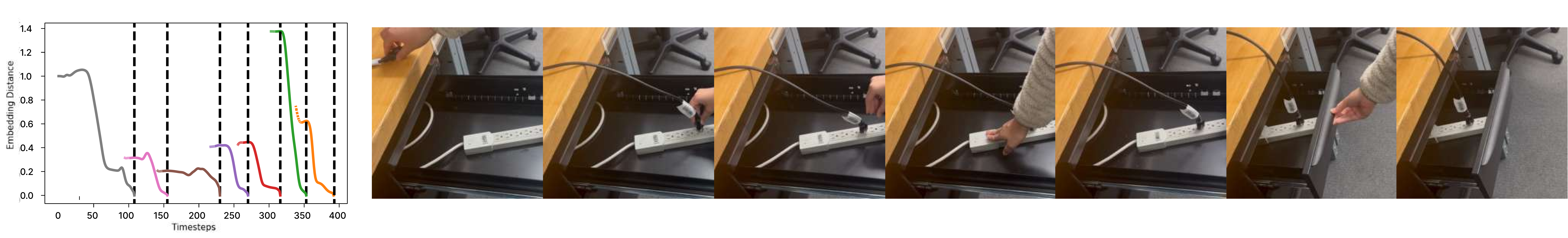}
    \caption{\textbf{Video sequence}: opening a drawer, picking the charger, plugging the charger, turning on the power strip, closing the drawer partially.}
    \label{appendix:fig:drawer}
\end{figure*}

\begin{figure*}
    \centering
    \includegraphics[width=1\textwidth]{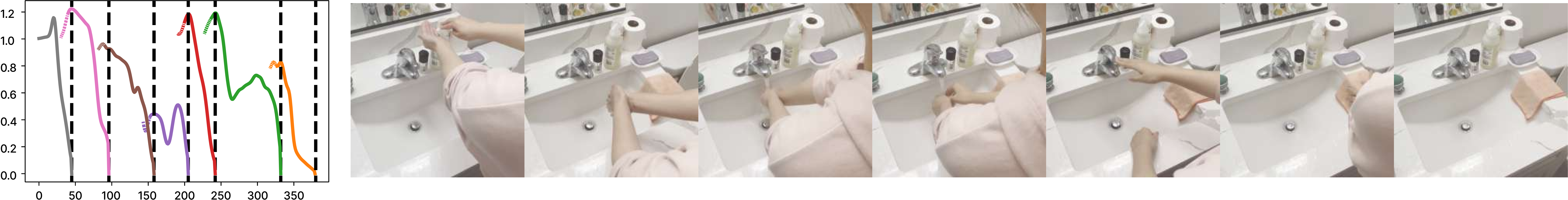}
    \caption{\textbf{Video sequence}: lathering hands, washing hands, turning off the tap, wiping hands with towel, placing back the cloth.}
    \label{appendix:fig:wash}
\end{figure*}

\begin{figure*}[htbp!]
    \centering
    \includegraphics[width=1\linewidth]{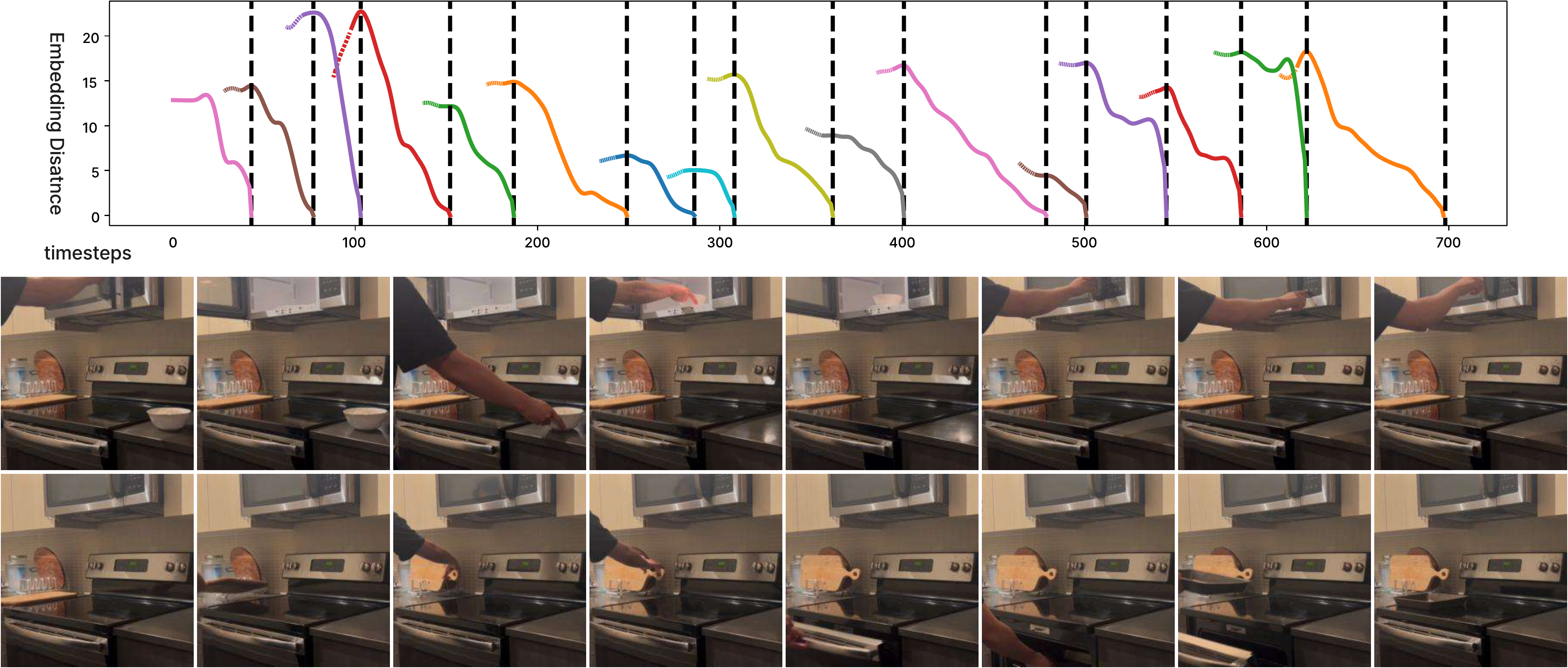}
    \caption{\textbf{UVD decomposes long video into subgoals.} The video sequence demonstrates: opening the microwave, placing a bowl with rice inside the microwave, closing the microwave, activating the microwave to heat the rice, placing the cutting board onto its rack, opening the oven, and putting the baking tray on the burner.}
    \label{appendix:fig:decomp-in-the-wild-longest}
\end{figure*}

\section{Limitation and Future Works}

While \acronym offers the advantage of not necessitating any task-specific knowledge or training, its efficacy is well-demonstrated across both simulated and real-robot environments. However, as we only validate on fully observable manipulation tasks, direct application to navigation tasks, especially those embodied tasks involving partial observations, may not yield intuitive or explainable subgoals (even though representations are pretrained with temporal objective using egocentric datasets~\cite{nair2022r3m, ma2022vip, ma2023liv}). 

Looking ahead, we are eager to broaden the applications of \acronym, diving deeper into its capabilities within egocentric scenarios, and even the key-frame extraction for video understanding and dense video caption tasks. On another front, while task graphs are widely used in Reset-Free RL~\cite{gupta2021reset, xu2023dexterous}, acquiring milestones as subgoals is resource-intensive and lacks scalability. By integrating our off-the-shelf \acronym subgoals into the task-graph, we are interested in seeing agents that, with minimal resets, can adeptly handle a wide range of tasks across various sequences and horizons in the wild.

\end{document}